\documentclass[10pt, a4paper]{article}

\usepackage{lrec-coling2024}

\usepackage{dirtytalk}
\usepackage{graphicx}
\usepackage{algorithm}
\usepackage{algpseudocode}
\usepackage{amsfonts}
\usepackage{booktabs}
\usepackage{multirow}
\usepackage{caption}
\usepackage{subcaption}

\usepackage{amsthm}
\theoremstyle{definition}
\newtheorem{definition}{Definition}[section]

\usepackage{amsmath}

\DeclareMathOperator*{\argmin}{argmin}

\algrenewcommand\algorithmicrequire{\textbf{Input:}}
\algrenewcommand\algorithmicensure{\textbf{Output:}}

\title{A Comparative Analysis of Word-Level Metric Differential Privacy: Benchmarking The Privacy-Utility Trade-off}

\name{Stephen Meisenbacher, Nihildev Nandakumar, \\ {\bf \large Alexandra Klymenko}, {\bf \large Florian Matthes}}

\address{Technical University of Munich \\ TUM School of Computation, Information and Technology \\ Department of Computer Science \\
         Garching, Germany \\
         \texttt{\small \{stephen.meisenbacher,nihildev.nandakumar,alexandra.klymenko,matthes\}@tum.de}\\}

\abstract{
The application of Differential Privacy to Natural Language Processing techniques has emerged in relevance in recent years, with an increasing number of studies published in established NLP outlets. In particular, the adaptation of Differential Privacy for use in NLP tasks has first focused on the \textit{word-level}, where calibrated noise is added to word embedding vectors to achieve \say{noisy} representations. To this end, several implementations have appeared in the literature, each presenting an alternative method of achieving word-level Differential Privacy. Although each of these includes its own evaluation, no comparative analysis has been performed to investigate the performance of such methods relative to each other. In this work, we conduct such an analysis, comparing seven different algorithms on two NLP tasks with varying hyperparameters, including the \textit{epsilon ($\varepsilon$)} parameter, or privacy budget. In addition, we provide an in-depth analysis of the results with a focus on the privacy-utility trade-off, as well as open-source our implementation code for further reproduction. As a result of our analysis, we give insight into the benefits and challenges of word-level Differential Privacy, and accordingly, we suggest concrete steps forward for the research field. 
 \\ \newline \Keywords{differential privacy, privacy-preserving NLP, evaluation} }

\begin{document}

\maketitleabstract

\thispagestyle{empty}

\section{Introduction}
Privacy vulnerabilities in Natural Language Processing (NLP) have recently been placed in the spotlight, and the discussions surrounding data privacy in this setting have gained increased attention with the rise of Large Language Models (LLMs) and chatbots such as ChatGPT. In particular, privacy risks have been demonstrated in embedding models \cite{song2020information,thomas2020investigating, morris2023text} and general-purpose language models \cite{9152761, carlini2021extracting}.

To combat privacy risks in data processing settings, Privacy-Enhancing Technologies (PETs) have emerged as concrete technical solutions that can be incorporated into existing systems. Under this class of technologies, Differential Privacy (DP) \cite{dwork2006differential} has risen in popularity due to its mathematical foundations, composability and robustness to post-processing, and above all, its flexible privacy parameter, known as $\varepsilon$. 

The application of DP to NLP settings does not come immediately, as the original sense of DP was designed for injecting plausible deniability into queries performed on sensitive attributes from structured databases. As textual data rarely exists in this form, reasoning about DP definitions initially comes with its challenges \cite{klymenko2022differential}. Nevertheless, a number of implementations have appeared in the literature, and as pointed out by Hu et al., the majority of these revolve around embedding vector perturbation methods at the word level \cite{hu2023differentially}. Many of these implementations employ \textit{Metric} Local Differential Privacy (MLDP), which was introduced as a generalization of the standard DP notion \cite{chatzikokolakis2013broadening}.

The focus on applying DP to word embeddings marks an intuitive first step in fusing the two fields, as words can be perceived as atomic units of information, which in turn are replaceable via calibrated perturbations. In such methods, the goal becomes to obfuscate the original text data as much as possible, while still preserving semantic coherence, and ideally, grammatical correctness. In terms of privacy preservation, several metrics are introduced in the literature, such as plausible deniability statistics \cite{feyisetan2020privacy} or membership inference attack performance \cite{shokri2017membership, carvalho2023tem}. Even so, such statistics are not uniformly reported across all word-level DP papers.

Beyond the metrics used to quantify the implications on privacy and utility, implementation papers do not run a standard evaluation, making a comparison in terms of performance quite difficult. The diversity in evaluation setups can be attributed both to the relative adolescence of the field and accordingly, the lack of a defined benchmark. 

In this work, we aim to address some of the above-mentioned gaps. We design an experimental setup with two separate NLP tasks, in which seven different word-level DP algorithms are tested. These experiments are run with various combinations of $\varepsilon$ and embedding dimension. Finally, a set of statistics is calculated on each experiment iteration, providing the foundation for a comparative analysis against the provided baselines.

The results from this work present the following contributions to the research of DP in NLP:
\begin{enumerate}
    \itemsep 0em
    \item An overview of the disparity in evaluation methods for word-level DP
    \item A novel multi-dimensional experimental setup focused on benchmarking privacy and utility metrics for word-level MLDP
    \item A comparative analysis of word-level MLDP methods, guided by a novel composite metric
    \item An open-source replication package for reproduction of the experiments, which includes previously unavailable code implementations of the selected methods, found at: \\ \url{https://github.com/sjmeis/MLDP}
\end{enumerate}

The structure of this paper is as follows. In Section \ref{sec:related}, related work in the field of word-level DP and its evaluation are discussed. Afterwards, in Section \ref{sec:foundations}, foundations of DP for NLP are introduced. Section \ref{sec:method} briefly outlines the followed methodology for this work, while Section \ref{sec:results} illustrates the resulting findings. These results are analyzed and discussed in Section \ref{sec:discuss}. Finally, Section \ref{sec:conclusion} underlines the implications following from our work and potential future directions, which is followed by a discussion of the perceived limitations of our study.

\section{Related Work}
\label{sec:related}
The investigation of Differential Privacy in Natural Language Processing, specifically on the word level, can be traced back to SynTF \cite{weggenmann2018syntf}, in which \say{synthetic} term-frequency vectors are created by performing single word replacements using the Exponential Mechanism \cite{mcsherry2007mechanism}. Fernandes presented the novel concept of using calibrated noise added directly to word embedding vectors to achieve noisy, perturbed vectors \cite{fernandes2019generalised}. This method relies on a generalized form of DP, often referred to as \textit{metric} DP, which relaxes DP for use in arbitrary vector spaces endowed with a metric \cite{chatzikokolakis2013broadening}. Further improvements to this technique were achieved by experimentation with underlying noise addition mechanisms, distance metrics, or both \cite{xu2020differentially, feyisetan2020privacy, carvalho2023tem}. These implementations focus on the \textit{local} DP setting \cite{kasiviswanathan2011can}, in which DP is applied to data at the user level and not at some central authority.

A recent survey \cite{hu2023differentially} categorizes DP-NLP methods into two categories: \textit{gradient perturbation} and \textit{vector embedding perturbation}. Of the 19 implementations listed under vector embedding perturbation in the local setting, 17 are word-level. 

\citet{klymenko2022differential} highlight the importance of benchmarking in DP-NLP, particularly as future research in the field. Looking to the word-level methods outlined by Hu et al., there is a great disparity in the tasks, datasets, and parameters used to evaluate the proposed methods. An overview of these evaluations, in line with the 17 mentioned methods, is provided in Table \ref{tab:eval}.

In the works presented in Table \ref{tab:eval}, utility is often measured by evaluating the accuracy of a given NLP task with perturbed input data. Privacy, on the other hand, is largely measured via (1) \textit{Empirical Privacy}, or the \textit{decrease} in performance for adversarial attacks, or  (2) \textit{Plausible Deniability}, in which statistics try to illustrate the level of plausible deniability introduced by a DP mechanism. Concretely, plausible deniability is often measured by estimating the probability that a word will be perturbed to another word, i.e., not remain the same.

Of particular focus in this work are the publications presented in the bottom half of Table \ref{tab:eval}. Specifically, we investigate MDP techniques in the local setting, henceforth Metric-LDP, or \textit{MLDP}. These methods generally focus on leveraging MLDP to add calibrated noise to static word embeddings (e.g., GloVe \cite{pennington-etal-2014-glove}), in order to achieve noisy word representations.

\begin{table*}[htbp]
\centering
\resizebox{\linewidth}{!}{
    \begin{tabular}{p{0.25\linewidth}|p{0.1\linewidth}|p{0.26\linewidth}|l|l|p{0.2\linewidth}|p{0.15\linewidth}}
    \toprule
    \textbf{Publication} & \textbf{Model} & \textbf{Task} & \textbf{Dataset} & \textbf{Epsilon} & \textbf{Privacy Metrics} & \textbf{Utility Metrics} \\ \hline
    \cite{lyu2} & BERT & \begin{tabular}[c]{@{}l@{}}Sentiment Analysis\\ Topic Classification\end{tabular} & \begin{tabular}[c]{@{}l@{}}Trustpilot\\ AG News/DW\end{tabular} & $\varepsilon \in \{0.05, 0.1, 0.5, 1, 5\}$ & Empirical Privacy & - \\
    \cite{lyu1} & \begin{tabular}[c]{@{}l@{}}GloVe\\ BERT\end{tabular} & \begin{tabular}[c]{@{}l@{}}Sentiment Analysis\\ Intent Detection\\ Paraphrase Identification\end{tabular} & \begin{tabular}[c]{@{}l@{}}IMDb, Yelp, Amazon\\ Intent Dataset\\ MRPC\end{tabular} & $\varepsilon \in \{0.5, 1, 5, 10\}$ & - & Accuracy, F1 \\
    \cite{plant} & BERT & Sentiment Analysis & Trustpilot & $\varepsilon \in \{0.01, 0.1, 0.5, 1\}$ & Empirical Privacy & Accuracy, F1 \\
    \cite{krishna} & LSTM & Intent Classification & ATIS, SNIPS & $\varepsilon \in \{0.25, 0.5, 0.6, 0.75, 0.85, 1\}$ & AUC & Accuracy \\
    \cite{habernal} & - & - & - & - & - & - \\
    \cite{igamberdiev} & LSTM & Intent Classification & ATIS, SNIPS & $\varepsilon \in \{1, 10, 100, 1000\}$ & - & F1 \\
    \cite{maheshwari} & Private Encoder & \begin{tabular}[c]{@{}l@{}}Sentiment Analysis\\ Attribute Detection\end{tabular} & \begin{tabular}[c]{@{}l@{}}Twitter\\ Bias in Bios, CelebA, Adult Income\end{tabular} & $\varepsilon \in \{8, 10, 12, 14, 16\}$ & Empirical Privacy, MDL & Accuracy \\
    \midrule
    \midrule
    \cite{feyisetan2020privacy} & \begin{tabular}[c]{@{}l@{}}GloVE\\ FastText\end{tabular} & \begin{tabular}[c]{@{}l@{}}Binary Classification\\ Multi-class Classification\\ Question Answering\end{tabular} & \begin{tabular}[c]{@{}l@{}}IMDb\\ Enron emails\\ InsuranceQA\end{tabular} & $\varepsilon \in \{6, 12, 17, 23, 29, 35, 41, 47, 52\}$ & Plausible Deniability & Accuracy \\
    \cite{xu2020differentially} & FastText & Binary Classification & Twitter, SMSSpam & $\varepsilon \in \{1, 5, 10, 20, 40\}$ & Plausible Deniability & Accuracy \\
    \cite{xu2} & \begin{tabular}[c]{@{}l@{}}GloVe\\ FastText\end{tabular} & \begin{tabular}[c]{@{}l@{}}Word Classification\\ Sentiment Analysis\\ Binary Classification\end{tabular} & \begin{tabular}[c]{@{}l@{}}Product Reviews\\ IMDb\\ Twitter\end{tabular} & $\varepsilon \in (0, 40]$ & Empirical Privacy & - \\
    \cite{xu3} & GloVe & \begin{tabular}[c]{@{}l@{}}Sentiment Analysis\\ Textual Entailment\end{tabular} & \begin{tabular}[c]{@{}l@{}}IMDb\\ SNLI\end{tabular} & N.A. & Empirical Privacy & Accuracy \\
    \cite{carvalho2023tem} & GloVe & Sentiment Analysis & IMDb & $\varepsilon \in (0, 22]$ & Empirical Privacy & Accuracy \\
    \cite{feyisetan2} & \begin{tabular}[c]{@{}l@{}}GloVe\\ FastText\end{tabular} & Various & MR, CR, MPQA, SST-5, TREC-6 & N.A. & - & Accuracy \\
    \cite{feyisetan3} & Poincaré & Various & \begin{tabular}[c]{@{}l@{}}MR, CR, MPQA, SST-5, TREC-6\\ SICK-E, MRPC, STS14\end{tabular} & $\varepsilon \in \{0.125, 0.5, 1, 2, 8\}$ & Plausible Deniability & Accuracy \\
    \cite{carvalho2} & Binary & Sentiment Analysis & IMDb & $\varepsilon \in (0, 22]$ & Empirical Privacy & Accuracy \\
    \cite{tang} & GloVe & \begin{tabular}[c]{@{}l@{}}Sentiment Analysis\\ Topic Classification\end{tabular} & \begin{tabular}[c]{@{}l@{}}Trustpilot\\ AG News\end{tabular} & $\varepsilon \in \{3, 4, 5, 6, 7, 8\}$ & Plausible Deniability & Accuracy \\
    \cite{yue} & \begin{tabular}[c]{@{}l@{}}GloVe\\ BERT\end{tabular} & \begin{tabular}[c]{@{}l@{}}Sentiment Analysis\\ Semantic Textual Similarity \\ Question Answering\end{tabular} & \begin{tabular}[c]{@{}l@{}}SST-2\\ MED-STS\\ QNLI\end{tabular} & $\varepsilon \in \{1, 2, 3\}$ & Empirical Privacy & Accuracy \\
    \bottomrule
    \end{tabular}
}
\caption{Word-level Local Differential Privacy (LDP) techniques and their evaluation. The bottom section denotes word-level \textit{metric} LDP approaches, the majority of which operate on static word embeddings.}
\label{tab:eval}
\end{table*}

In another recent work \cite{mattern2022limits}, the authors address potential shortcomings of word-level DP, particularly the tight constraints placed in the local DP setting, as well as the effect this has on the quality of language output, leading to grammatical errors and inflexibility when attempting to enforce changes in syntax. The implications of these findings on model performance were not discussed or analyzed, however, and it is here where our investigation begins.

\section{Foundations}
\label{sec:foundations}
Differential Privacy \cite{dwork2006differential} was introduced in 2006 as a formal definition for the quantification of individual privacy. The original notion as proposed by Dwork was aimed at privacy preservation in the centralized setting, in which each row of a structured database corresponds to one individual's data. According to Differential Privacy, the inclusion of an individual in the dataset should only affect the outcome of aggregate queries by a certain bound, governed by the $\varepsilon$ (privacy) parameter.

\begin{definition}[$\varepsilon$-Differential Privacy]
For any databases $D_1$ and $D_2$ differing in exactly one element, any $\varepsilon > 0$, a randomized function $\mathcal{K}$, and all $\mathcal{S} \subseteq Range(\mathcal{K})$:
\begin{equation}
    Pr[\mathcal{K}(D_1) \in \mathcal{S}] \le e^\varepsilon Pr[\mathcal{K}(D_2) \in \mathcal{S}]
\label{eq:DP}
\end{equation}
\end{definition}
Thus, the $\varepsilon$ parameter determines how \textit{indistinguishable} the output (distribution) of the operation performed on $D_1$ and $D_2$ must be.

As mentioned, the focus of this work is placed on the application of DP in the word embedding space, which is a multi-dimensional \textit{vector space}. As such, the definition of Equation \ref{eq:DP} is not readily transferable to this space. Instead, the notion of MDP was developed to incorporate the usage of a distance metric within the word vector space.

\begin{definition}[Metric Differential Privacy, or $d_\mathcal{X}$-privacy]
For any $x, x' \in \mathcal{X}$ (vector space) endowed with a metric $d$, any $\varepsilon > 0$, and a randomized function $M: \mathcal{X} \rightarrow \mathcal{Y}$:
\begin{equation}
Pr[M(x) \in \mathcal{Y}] \le e^{\varepsilon d(x,x')} Pr[M(x') \in \mathcal{Y}]
\label{eq:MDP}
\end{equation}
\end{definition}
One can see that this definition incorporates the metric $d$ into the $\varepsilon$ parameter, now scaling the required indistinguishability by the relationship between two inputs (i.e., two words from the vocabulary $\mathcal{X}$). In this setting, one can now reason about two word vectors, whose relation can be quantified by a distance metric.

In order to define MDP in the local setting, the notion of MLDP  was introduced \cite{alvim2018local}, which is defined below:

\begin{definition}[Metric Local Differential Privacy]
For all $y \in \mathcal{Y}$:
\begin{equation}
    Pr[M(x) = y] \le e^{\varepsilon d(x,x')} Pr[M(x') = y]
\label{eq:LMDP}
\end{equation}
\end{definition}

For an in-depth introduction of DP in metric spaces for NLP, we refer the reader to \cite{feyisetan2020privacy}. For technical details on how calibrated noise is generated using a variety of DP mechanisms, we refer to \cite{barthe2016programming}.

\section{Methodology}
\label{sec:method}
In this section, we introduce the details of our experimental design, as well as provide a brief overview of the algorithms included in our analysis.

\subsection{Experimental Design}
\subsubsection{Tasks and Datasets}
For our comparative analysis, we benchmark the selected methods on two NLP tasks: Sentiment Analysis and Topic Classification. 

The Sentiment Analysis task is run on the IMDb Movie Review Dataset \cite{maas-EtAl:2011:ACL-HLT2011}, which is a dataset of 50k movie reviews, classified as either negative or positive in sentiment. We take a random sample of 12k movies -- 8k for training, 2k for validation, and 2k for testing.

AG News \cite{Zhang2015CharacterlevelCN} is a dataset of nearly 130k text excerpts from AG. The dataset contains news articles on the four largest topics in the AG News corpus: world, sports, business, and science. For this task, we take a random sample of 6k articles from each topic -- 16k in total for training, 4k for validation, and 4k for testing.

A summary of both selected tasks and their underlying datasets is included in Table \ref{tab:summary}.

\begin{table}[htbp]
\small
\centering
\begin{tabular}{@{}lcc@{}}
\toprule
Dataset & \textbf{IMDb} & \textbf{AG News} \\ 
\midrule
Task Type & Binary & Multi-class \\
Training set size & 10,000 & 20,000 \\
Test set size & 2,000 & 4,000 \\
Total word count & 808,382 & 510,582 \\
Vocabulary Size & 42,662 & 27,234 \\
\multirow{2}{*}{Sentence Length} & µ = 80.84 & µ = 25.52 \\
 & $\sigma$ = 22.53 & $\sigma$  = 6.78 \\
\bottomrule
\end{tabular}
\caption{Summary of the two selected NLP tasks and datasets, with key characteristics.}
\label{tab:summary}
\vspace{-10pt}
\end{table}

\paragraph{Choice of Datasets}
We choose IMDb and AG News for multiple reasons: (1) their utilization in previous works \cite{tang, carvalho2023tem, xu2, feyisetan2020privacy}, (2) their relatively large size and accessibility, (3) the ability of IMDb to simulate \say{sensitive} information (personal reviews), and (4) the multi-class classification problem of AG News, to supplement the simpler binary case of IMDb.

\subsubsection{Evaluation Model}
For both tasks, we employ an LSTM-based model \cite{hochreiter1997long} using Keras. An embedding layer is added to allow the use of GloVe embeddings as input, followed by a Dropout layer (0.2), and finally, a fully connected layer with a softmax activation is added to facilitate both classification tasks. The embedding layer was added to facilitate the input format, as all tested mechanisms map input words to discrete \say{noisy} output words, each of which corresponds to an embedding (GloVe in this case). For each experiment configuration, the model was trained with a batch size of 64 with a maximum of 30 epochs. Early stopping and checkpointing were activated. 

It should be noted that the choice of LSTM in lieu of transformer-based models was justified for two reasons: (1) as the nature of this study is to benchmark against a baseline, it was not seen as necessary to achieve SOTA performance on the two chosen tasks, and relatedly, (2) the training of LSTMs is far more efficient than that of transformer-based models, and given the large dimensions of our conducted evaluation, the LSTM provided the much more time- and resource-efficient option.

For each experimental setting, the MLDP perturbed data (train + validation split) is used for training, and the evaluation is performed on the trained model using the test split perturbed by the same mechanism. This is to simulate the local DP setting, in which user data is perturbed locally. The metrics of the models trained on perturbed data were captured and compared against the non-DP (original data) baseline.


\subsubsection{Embedding Model}
We utilize pre-trained GloVe embeddings \cite{pennington-etal-2014-glove}, which were trained on the entire Wikipedia dataset from 2014 and Gigaword 5\footnote{\url{https://nlp.stanford.edu/projects/glove/}}. In particular, we use the 50-, 100-, and 300-dimension versions of the word embedding model.

\subsubsection{Privacy Budget}
As introduced in Section \ref{sec:foundations}, the privacy parameter known as epsilon ($\varepsilon$) is used to scale the level of privacy desired. In practice, a smaller epsilon value adds a higher level of noise and thus leads to a higher theoretical level of privacy protection. As a result of surveying the parameter choice of the MLDP methods in Table \ref{tab:eval}, we choose to use $\varepsilon \in \{1, 5, 10\}$ for our experiments.

\subsubsection{Metrics}
\label{sec:metrics}
For utility, we report the accuracy of each experiment. This is seen to be a sufficient indicator of performance, as both tasks have balanced datasets.

For privacy estimation, we employ four metrics which allow for a uniform platform for evaluating the effect of each algorithm on the input text. The first three methods, introduced below, were chosen due to their usage in the previous works listed in Table \ref{tab:eval}, while the final metric (LOW) is inspired by the approach of \citeauthor{yue}.

\paragraph{Plausible Deniability (PD) [$N_w \downarrow, S_w \uparrow$]} These two metrics measure the \textit{plausible deniability} created by a particular mechanism. Concretely, given a word $n$, $N_w$ estimates the probability of not modifying the word (word stays unperturbed), and $S_w$ measures the support of the set of output words, i.e., the number of words that $w$ can be perturbed with probability $1 - \eta$, with $\eta$ being small. Inspired by the approach taken by \cite{feyisetan2020privacy, xu2}, we estimate these metrics by running each mechanism on a set of 25 random words from each dataset, each word being run 100 times.

\paragraph{Perturbation Percentage (PP) $\uparrow$} This metric calculates the percentage of words perturbed to a different word, i.e., that did not remain unchanged. The more perturbed the data is, the higher the privacy protection; however, this may come at the expense of reduced data utility \cite{kang2020input}.

\paragraph{Cosine Similarity (CS) $\uparrow$} This metric calculates the cosine similarity between the sentence embedding representations of the original and perturbed inputs, inspired by BERTScore \cite{zhang2019bertscore}. Although not a privacy metric per se, cosine similarity provides insight into the effect of word-level perturbations. By using this metric in combination with PP, one can gauge the trade-off between input perturbation and preservation of meaning. To calculate this metric, a pre-trained SBERT model \cite{reimers-2019-sentence-bert} is used, namely \textbf{paraphrase-distilroberta-base-v1}.

\paragraph{Least-Occuring Words (LOW) $\downarrow$} This metric calculates the percentage of least-occurring words existing in both the original and perturbed datasets. Least-occurring words are considered sensitive because they may contain information about individuals, potentially leading to privacy breaches \cite{yue}. Here, we calculate the percentage of the 1000 least-occurring words from the original dataset that are still present in the perturbed data.

\subsection{Selected Algorithms}
\label{ssec:selected}
We introduce the seven methods included in this study, which represent all existing word-level MLDP approaches that operate on static word embeddings in the Euclidean space. In addition, we also include one method (SynTF) as a baseline, as it served as a precursor to all other evaluated methods. Finally, we also briefly discuss the excluded methods from Table \ref{tab:eval}.

\paragraph{SynTF \cite{weggenmann2018syntf}}
Although not explicitly an MLDP method, the SynTF mechanism can be viewed as a precursor, as it performs word-level DP synonym replacements by sampling words from term-frequency vectors.

\paragraph{Calibrated Multivariate Perturbations (CMP) \cite{feyisetan2020privacy}} This method adds calibrated multivariate normal noise to word embeddings, and then perturbs the noisy vectors back to the nearest neighbor in the embedding space.

\paragraph{Mahalanobis Mechanism \cite{xu2020differentially}} This method aims to improve upon previous methods by adding elliptical noise using the regularized Mahalanobis norm, in order to account for vectors existing in sparse regions of the embedding space.

\paragraph{SanText \cite{yue}} The SanText mechanism aims to improve word perturbation by relating the perturbation probability of a word to another token by their Euclidean distance in the embedding space. Thus, the closer two words are semantically, the higher the probability that one serves as the replacement for the other. We use the base mechanism proposed in the paper.

\paragraph{Truncated Gumbel Mechanism \cite{xu3}} This method utilizes calibrated Gumbel noise to scale the probability of perturbing to a new word within a selected set of candidate words.

\paragraph{Vickrey Mechanism \cite{xu2}} This mechanism, motivated by Vickrey auctions, balances the perturbation probability between the first and second nearest word neighbors. The authors also provide a generalized mechanism for $k$ neighbors; we implement the original method (k=2).

\paragraph{Truncated Exponential Mechanism (TEM) \cite{carvalho2023tem}} This mechanism generalizes the perturbation process to a \textit{selection problem} by utilizing the Exponential Mechanism. For our study, we utilize the TEM mechanism with Euclidean distance, as proposed in the paper.

\subsubsection{Excluded Algorithms}
Of the introduced MLDP methods presented in Table \ref{tab:eval}, we exclude \cite{feyisetan3} and \cite{carvalho2} due to their use of embeddings in non-euclidean spaces. Similarly, we exclude \cite{feyisetan2}, as this method does not map noisy vectors to words. \cite{tang} is excluded due to its multi-stage perturbation mechanism and its similarity to CMP.

\subsubsection{Algorithm-specific Parameters}
\label{sec:params}
In our experiments, we used the following algorithm-specific parameters (beyond $\epsilon$), which can be found and modified in our provided repository: \textbf{SynTF}: synonyms from NLTK WordNet, \textbf {Mahalanobis}: $\lambda = 0.2$, \textbf{Vickrey}: $t = 0.5$, \textbf{TEM}: $\gamma = 0.5$.

\begin{table*}[htbp]
\resizebox{\linewidth}{!}{
\begin{tabular}{l|lllllllll||lllllllll}
\toprule
\multicolumn{1}{r|}{Task:}                      & \multicolumn{9}{c||}{\textbf{Sentiment Analysis (IMDb)}}                                                               & \multicolumn{9}{c}{\textbf{Topic Classification (AG News)}}                                                            \\ \cline{2-19} 
\multicolumn{1}{r|}{Baseline:}  & \multicolumn{3}{c|}{77.30}& \multicolumn{3}{c|}{79.81}  & \multicolumn{3}{c||}{84.53} & \multicolumn{3}{c|}{83.92}   & \multicolumn{3}{c|}{84.44}  & \multicolumn{3}{c}{84.28} \\ \cline{2-19} 
\multicolumn{1}{r|}{Dimension:}  & \multicolumn{3}{c|}{50}           & \multicolumn{3}{c|}{100}          & \multicolumn{3}{c||}{300} & \multicolumn{3}{c|}{50}           & \multicolumn{3}{c|}{100}  & \multicolumn{3}{c}{300} \\ \cline{2-19} 
\multicolumn{1}{r|}{Epsilon:} & 1  & 5  & \multicolumn{1}{l|}{10} & 1  & 5  & \multicolumn{1}{l|}{10} & 1      & 5      & 10     & 1  & 5  & \multicolumn{1}{l|}{10} & 1  & 5  & \multicolumn{1}{l|}{10} & 1  & 5  & 10    \\ 
\midrule
\midrule
SynTF       & 72.58 & 72.26 & \multicolumn{1}{l|}{71.83} & 74.16 & 74.79 &  \multicolumn{1}{l|}{74.64}           & 75.96    & 75.99   & 76.24  & 81.73 & 81.55 & \multicolumn{1}{l|}{81.05} & 84.36 & 82.53 & \multicolumn{1}{l|}{81.71} & \textbf{83.69}     & 82.59     & 82.39    \\
CMP         & 52.10 & 61.50 & \multicolumn{1}{l|}{\textbf{\underline{77.40}}} & 49.60 & 63.15 & \multicolumn{1}{l|}{78.85}           & 50.85   & 56.80  & 75.00  & 63.67 & 83.77 & \multicolumn{1}{l|}{\underline{84.60}} & 58.70 & 80.35 & \multicolumn{1}{l|}{84.26} & 51.30     & 69.55     & 82.45    \\
Mahalanobis & 53.20 & 66.70 & \multicolumn{1}{l|}{75.25} & 49.80 & 58.95 & \multicolumn{1}{l|}{76.25}            & 52.00    & 54.05   & 67.90  & 62.67 & 79.94 & \multicolumn{1}{l|}{\underline{84.32}} & 62.75 & 79.42 & \multicolumn{1}{l|}{83.70} & 53.05     & 68.32     & \textbf{\underline{85.75}}    \\
SanText     & 58.97 & 54.60 & \multicolumn{1}{l|}{54.52} & 64.48 & 63.16 & \multicolumn{1}{l|}{61.49}            & 71.21    & 68.94   & 69.38  & 65.15 & 63.91 & \multicolumn{1}{l|}{62.46} & 68.21 & 69.62 & \multicolumn{1}{l|}{68.98} & 73.20     & 74.08     & 74.11    \\
Gumbel      &  \textbf{\underline{78.08}}  &  \textbf{76.95} & \multicolumn{1}{l|}{\underline{77.33}} &  \textbf{78.43} &  \textbf{79.10} & \multicolumn{1}{l|}{\textbf{\underline{81.01}}}  & \textbf{81.64} &  81.08  &  81.24 & \textbf{\underline{84.17}} & \underline{83.95} & \multicolumn{1}{l|}{\underline{84.98}}   &  \textbf{\underline{85.27}}  &  \textbf{\underline{84.68}} & \multicolumn{1}{l|}{84.00} & 83.45 &  \textbf{\underline{84.80}} &  \underline{84.69} \\
Vickrey     &  54.18  &  70.00  & \multicolumn{1}{l|}{75.16}   &  50.50  &  69.46  & \multicolumn{1}{l|}{74.34}  & 53.66 &  59.00  & 69.23  &  63.29  &  81.00  & \multicolumn{1}{l|}{83.41}   &  59.78  &  76.95 & \multicolumn{1}{l|}{83.97} & 46.29   &   69.70   &   81.94    \\ 
TEM         & 52.38 & 74.50 & \multicolumn{1}{l|}{76.30} & 55.15 & 77.50 & \multicolumn{1}{l|}{78.75}            & 50.34    & \textbf{81.90} & \textbf{83.20}  & 65.15 & \textbf{\underline{84.70}} & \multicolumn{1}{l|}{\textbf{\underline{85.10}}} & 62.42 & 84.25 & \multicolumn{1}{l|}{\textbf{\underline{85.20}}} & 60.12     & 84.00     & 81.75    \\
\bottomrule
\end{tabular}
}
\caption{Utility scores (accuracy) for all experimental settings. The scores represent an average of five runs (10 for baseline). Bolded scores denote the highest score per setting, while underlined scores mark those that surpass the baseline.}
\label{tab:utility}
\end{table*}

\begin{figure*}[ht!]
    \centering
    \begin{subfigure}[htbp]{0.32\linewidth}
        \centering
        \includegraphics[width=\textwidth]{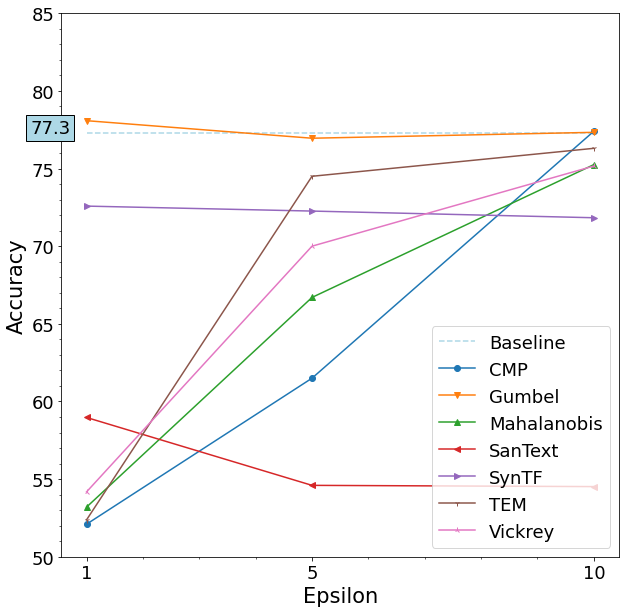}
        \vspace{-15pt}
        \caption{IMDb, \textit{d}=50}
        \label{fig:imdb50}
        \vspace{5pt}
    \end{subfigure}
    \hfill
    \begin{subfigure}[htbp]{0.32\linewidth}
        \centering
        \includegraphics[width=\textwidth]{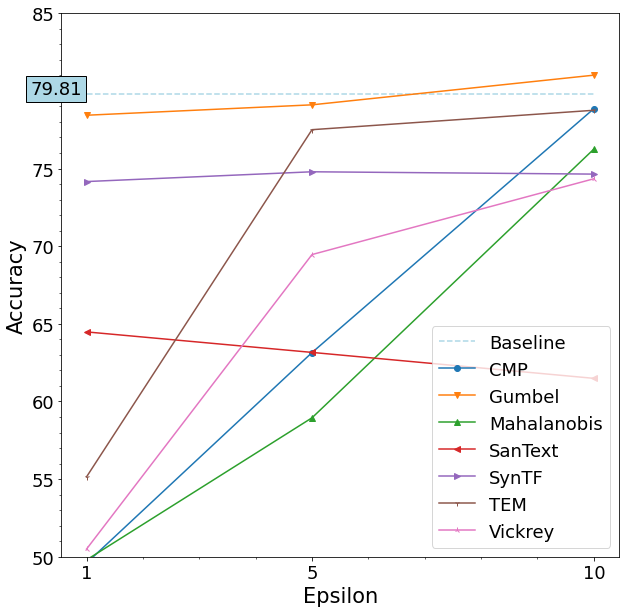}
        \vspace{-15pt}
        \caption{IMDb, \textit{d}=100}
        \label{fig:imdb100}
        \vspace{5pt}
    \end{subfigure}
    \hfill
    \begin{subfigure}[htbp]{0.32\linewidth}
        \centering
        \includegraphics[width=\textwidth]{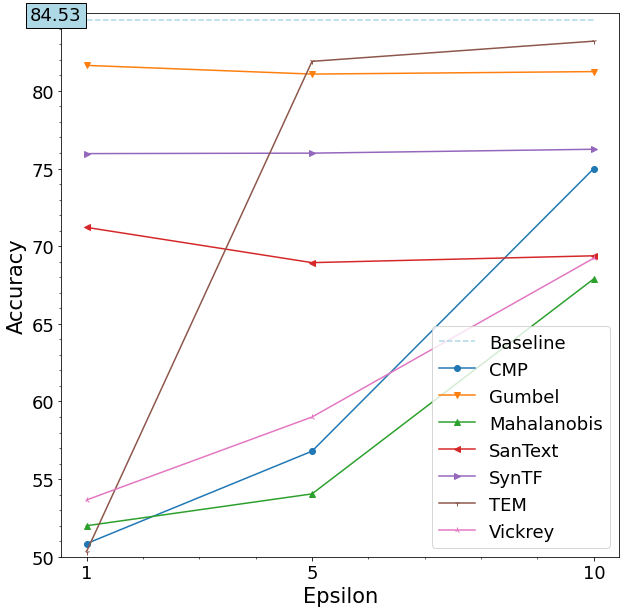}
        \vspace{-15pt}
        \caption{IMDb, \textit{d}=300}
        \label{fig:imdb300}
        \vspace{5pt}
    \end{subfigure}
    \begin{subfigure}[htbp]{0.32\linewidth}
        \centering
        \includegraphics[width=\textwidth]{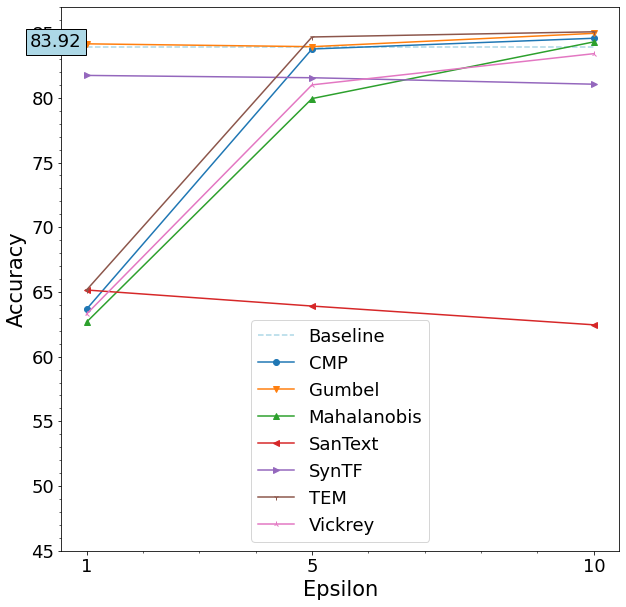}
        \textbf{\vspace{-15pt}}
        \caption{AG News, \textit{d}=50}
        \label{fig:ag50}
    \end{subfigure}
    \hfill
    \begin{subfigure}[htbp]{0.32\linewidth}
        \centering
        \includegraphics[width=\textwidth]{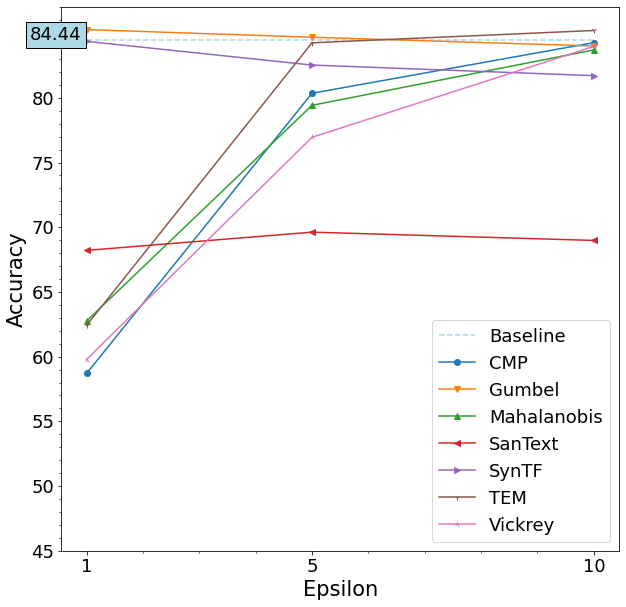}
        \vspace{-15pt}
        \caption{AG News, \textit{d}=100}
        \label{fig:ag100}
    \end{subfigure}
    \hfill
    \begin{subfigure}[htbp]{0.32\linewidth}
        \centering
        \includegraphics[width=\textwidth]{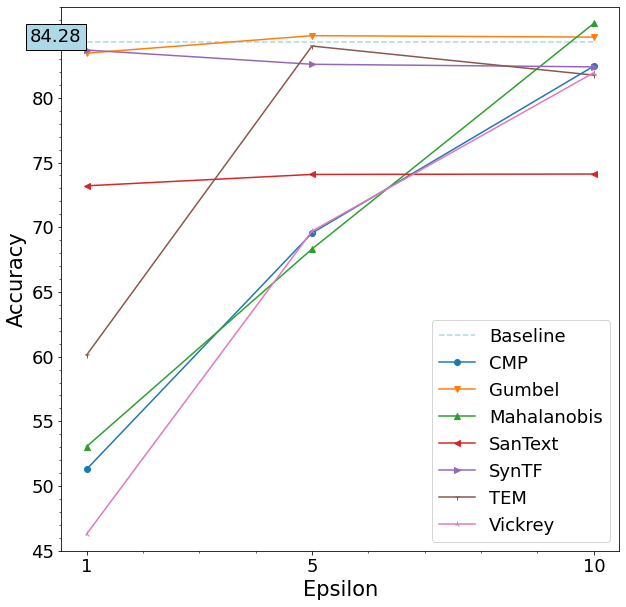}
        \vspace{-15pt}
        \caption{AG News, \textit{d}=300}
        \label{fig:ag300}
    \end{subfigure}
    \caption{Accuracy scores per task and embedding dimension (\textit{d}). Baseline scores are marked with a dotted line, and the baseline value is indicated in the light blue box. The scale of the y-axis is uniform across sub-figures for comparability.}
    \label{fig:acc_chart}
\end{figure*}

\section{Experiment Results}
\label{sec:results}
\subsection{Utility}
The utility results for our study are presented in Table \ref{tab:utility}. To obtain the accuracy scores, the LSTM model was trained five times (10 for baseline tests), and the evaluated scores were averaged to achieve a single score. As early stopping was implemented, accuracy was drawn from the best model. For each combination of (\textit{task}, \textit{dimension}, \textit{epsilon}), the highest score is \textbf{bolded}. Scores that surpass their respective (\textit{task}, \textit{dimension}, \textit{epsilon}) benchmark are \underline{underlined}. The measured scores are broken down by embedding dimension and task in Figure \ref{fig:acc_chart}.

\subsection{Privacy}
\begin{figure*}[htbp]
    \centering
    \begin{subfigure}[htbp]{0.49\textwidth}
        \centering
        \includegraphics[scale=0.4]{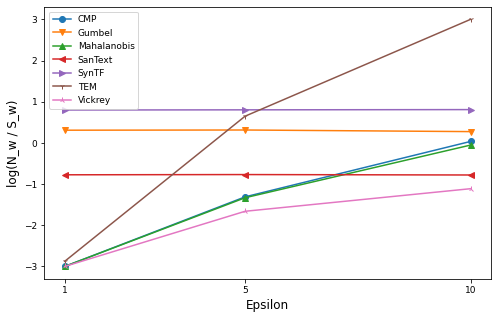}
        \vspace{-5pt}
        \caption{IMDb, $N_w$:$S_w$ ratio}
        \label{fig:imdb_ns}
    \end{subfigure}
    \hfill
    \begin{subfigure}[htbp]{0.49\textwidth}
        \centering
        \includegraphics[scale=0.4]{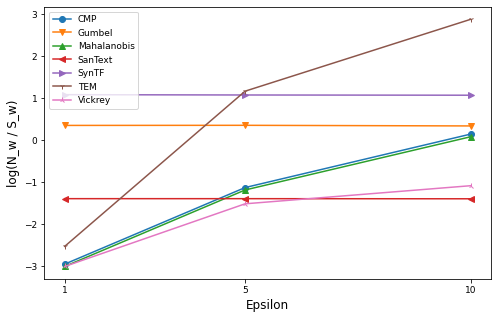}
        \vspace{-5pt}
        \caption{AG News, $N_w$:$S_w$ ratio}
        \label{fig:ag_ns}
    \end{subfigure}
    \caption{Ratio of $N_w$ to $S_w$, averaged over three embedding dimensions. Lower ratios correspond to higher plausible deniability. Ratios are shown on the logarithmic scale due to outlier values (i.e., for TEM).}
    \label{fig:ns_ratio}
\end{figure*}

For readability, we present summarizing visualizations of the privacy metric results. 
Figure \ref{fig:ns_ratio} illustrates the $N_w$:$S_w$ ratio of the selected methods for our three chosen $\epsilon$ values. As lower $N_w$ and higher $S_w$ values are preferred, lower values on the graph represent higher plausible deniability guarantees. 

\paragraph{A New Composite Metric}
The study of text privatization, for example in the case of word-level MLDP, often views the privacy-utility trade-off in two separate lights: privacy and utility are measured separately, and then these results are fused in a qualitative analysis. As such, to the best of the authors' knowledge, there exists no single metric that compares privacy and utility \textit{simultaneously}. We aim to address this gap in the introduction of the following metric.

In order to aid in our pursuit to benchmark the privacy-utility trade-off for our selected MLDP methods, we introduce a new metric that aims to capture both the utility of a method and its privacy-preserving capabilities. As such, we utilize the Privacy-Utility Composite (PUC) score, defined as:
\begin{equation}
    \resizebox{0.95\linewidth}{!}{$
    PUC = \alpha(\frac{100 * Acc}{B-Acc}) + (1-\alpha) \frac{(100-N_w) + S_w + PP + CS + (100-LOW)}{5}
    $}
\end{equation}
Where \textit{(B-)Acc} represents the (baseline) accuracy percentage, and $\{N_w, S_w, PP, CS, LOW\}$ represents the set of privacy metrics we use, as introduced in Section \ref{sec:metrics}. Scores where lower values are better are subtracted from 100. $\alpha$ is the \textit{privacy-utility tuning parameter}, where one can scale the preferred weight placed upon utility and privacy metrics. For example, choosing an $\alpha$ of 0.75 would strongly emphasize the weight of the utility outcomes, whereas the composite privacy score would receive a weight of only 0.25. More generally, we can define the PUC score as follows:
\begin{definition}[Privacy-Utility Composite (PUC $\uparrow$) Score]
For a set of (1) utility metrics $\mathcal{U}$, representing percentages compared to a baseline, and (2) a set of privacy metrics $\mathcal{P}$, which can be broken down into metrics $\mathcal{P}\uparrow$ and $\mathcal{P}\downarrow$, and a privacy-utility tuning parameter $\alpha$, and a max score $M$:
\begin{equation}
    \resizebox{0.95\linewidth}{!}{$
    PUC = \frac{\alpha}{|\mathcal{U}|} \sum_{u \in \mathcal{U}} u_i + \frac{1 - \alpha}{|\mathcal{P}|}\Biggl(\sum_{p \in \mathcal{P}\uparrow}p_i + \sum_{\tilde{p} \in \mathcal{P}\downarrow}(M-\tilde{p}_i)\Biggr)
    $}
\label{eq:PUC}
\end{equation}
\end{definition}

Note that Equation \ref{eq:PUC} assumes that all scores are on the same scale, i.e., in the range of $[0, M]$. Scores not on the same scale can be scaled or normalized accordingly. The PUC score also assumes an equal weighting within a metric set, e.g., all utility scores are weighted equally.

It should also be noted that the choice if $\alpha$ is envisioned to be performed \textit{a priori}, or rather, before the design or evaluation of MLDP mechanisms and not as a tunable parameter. In this way, the requirement of privacy protection versus utility preservation should be decided upon beforehand, so that such a balance will be reflected in the analysis of the PUC scoring results.

\begin{figure*}[ht!]
    \centering
    \begin{subfigure}[htbp]{0.32\textwidth}
        \centering
        \includegraphics[width=\textwidth]{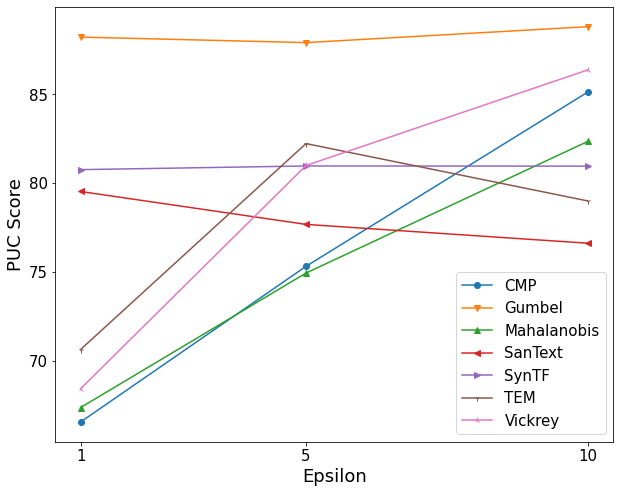}
        \vspace{-15pt}
        \caption{IMDB, $\alpha=0.75$}
        \label{fig:pucimdb75}
        \vspace{5pt}
    \end{subfigure}
    \hfill
    \begin{subfigure}[htbp]{0.32\textwidth}
        \centering
        \includegraphics[width=\textwidth]{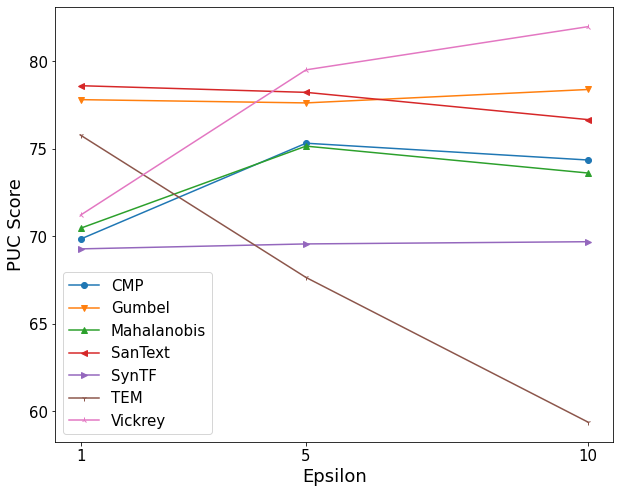}
        \vspace{-15pt}
        \caption{IMDB, $\alpha=0.5$}
        \label{fig:pucimdb5}
        \vspace{5pt}
    \end{subfigure}
    \hfill
    \begin{subfigure}[htbp]{0.32\textwidth}
        \centering
        \includegraphics[width=\textwidth]{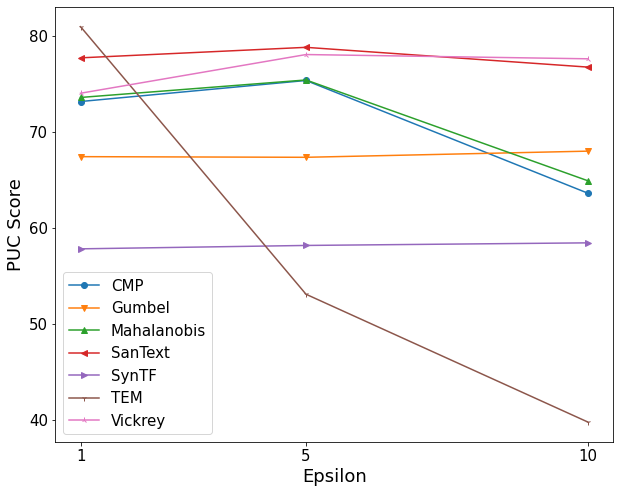}
        \vspace{-15pt}
        \caption{IMDB, $\alpha=0.25$}
        \label{fig:pucimdb25}
        \vspace{5pt}
    \end{subfigure}
    \begin{subfigure}[htbp]{0.32\textwidth}
        \centering
        \includegraphics[width=\textwidth]{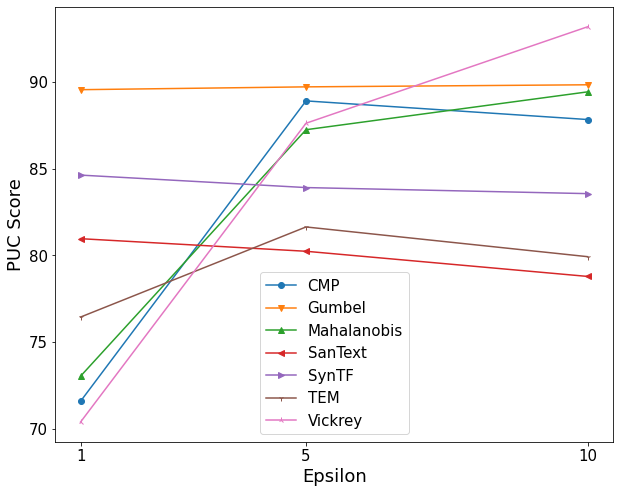}
        \vspace{-15pt}
        \caption{AG News, $\alpha=0.75$}
        \label{fig:pucag75}
    \end{subfigure}
    \hfill
    \begin{subfigure}[htbp]{0.32\textwidth}
        \centering
        \includegraphics[width=\textwidth]{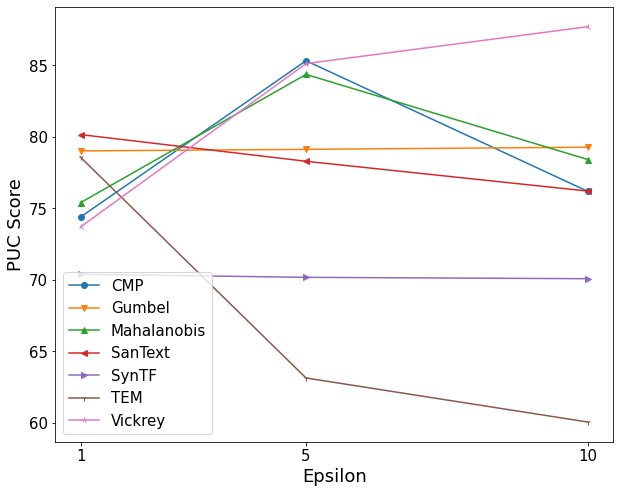}
        \vspace{-15pt}
        \caption{AG News, $\alpha=0.5$}
        \label{fig:pucag5}
    \end{subfigure}
    \hfill
    \begin{subfigure}[htbp]{0.32\textwidth}
        \centering
        \includegraphics[width=\textwidth]{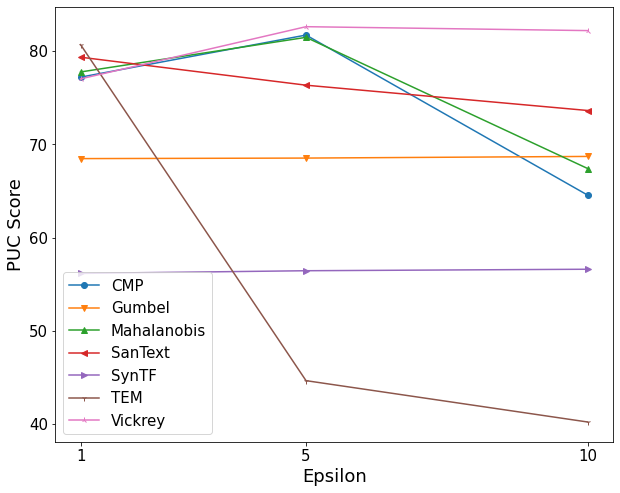}
        \vspace{-15pt}
        \caption{AG News, $\alpha=0.25$}
        \label{fig:pucag25}
    \end{subfigure}
    \caption{Privacy-Utility Composite (PUC) scores per task, with varying $\alpha$. The PUC scores were averaged across embedding dimension, and these averages are shown for each epsilon ($\epsilon$) value. The left column with $\alpha=0.75$ favors utility, the middle with $\alpha=0.5$ is balanced, and the right with $\alpha=0.25$ favors privacy.}
    \label{fig:puc_chart}
\end{figure*}

In Figure \ref{fig:puc_chart}, we illustrate average PUC scores with three $\alpha$ values. These values are meant to simulate a preference for utility (Figures \ref{fig:pucimdb75}, \ref{fig:pucag75}), a balanced preference (Figures \ref{fig:pucimdb5}, \ref{fig:pucag5}), and a preference for privacy preservation (Figures \ref{fig:pucimdb25}, \ref{fig:pucag25}).

Finally, to explore the relevance of our \textit{CS} privacy metric, we perform a Multiple Linear Regression (MLR) test. We use epsilon ($\epsilon$), $N_w$, $S_w$, and \textit{PP} as our predictor variables and \textit{CS} as our response variable. The resulting model and summary of the regression test are shown in Table \ref{tab:mlr}. Most importantly, one can see strong correlations between \textit{CS} and all other variables, notably a strong positive correlation with $\varepsilon$ and a strong negative correlation with \textit{PP}. This demonstrates that the choice of $\varepsilon$ is a strong indicator of the expected degree of utility in the output privatized text.

\begin{table}[htbp]
\small
\resizebox{0.95\linewidth}{!}{
\begin{tabular}{r|llll}
\toprule
\multicolumn{1}{l}{$R^2=0.905$} & \multicolumn{1}{c}{\textbf{coef.}} & \multicolumn{1}{c}{\textbf{std err}} & \multicolumn{1}{c}{\textbf{t}} & \multicolumn{1}{c}{\textbf{P>|t|}} \\
\midrule
\midrule
const. & 126.6834 & 4.963 & 25.528 & 0.000 \\
epsilon ($\varepsilon$) & 0.6732 & 0.192 & 3.503 & 0.001 \\
$N_w$ & -0.3158 & 0.054 & -5.818 & 0.000 \\
$S_W$ & -0.2157 & 0.028 & -7.811 & 0.000 \\
PP & -0.7728 & 0.038 & -20.337 & 0.000 \\
\bottomrule
\end{tabular}
}
\caption{MLR to predict the \textit{CS} metric. In general, $R^2$ measures the goodness of the fit, ranging from 0 to 1. All predictors are statistically significant.}
\label{tab:mlr}
\vspace{-10pt}
\end{table}

\section{Discussion}
\label{sec:discuss}
\paragraph{Effect of DP on Utility}
While it is reasonable that a lower $\epsilon$ value will generally lead to lower utility, a thorough study of our results reveals additional insights. Firstly, some methods prove to be \textit{utility loss invariant} w.r.t. the choice of $\epsilon$. With SynTF, this is explainable by its mechanism design. However, in the case of Gumbel and SanText, their strength in preserving utility across $\epsilon$ values is made clear.

We also observe the effect of embedding dimension. Looking at Figure \ref{fig:acc_chart}, one can see that as embedding dimension increases, Mahalanobis, CMP, and Vickrey all experience drops in accuracy, given a fixed $\epsilon$. It is presumed that this is an artifact of the mechanism design, where utility may begin to deteriorate as more dimensions of noise are added.

Notably, the use of an MLDP mechanism in some cases actually contributes to an increase in accuracy against the baseline, particularly in the case of Gumbel, but also observed in TEM, CMP, and Mahalanobis. Such a phenomenon was observed 13 times in the Topic Classification task and four times in Sentiment Analysis. This finding opens the discussion of MLDP as a \say{robustness mechanism}.

\paragraph{Privacy Analysis}
Moving to the analysis of our privacy metrics, we begin with our PD statistics, namely the ratio of $N_w$ to $S_w$. Much like Figure \ref{fig:acc_chart}, one can observe in Figure \ref{fig:ns_ratio} that three methods maintain a near-constant average ratio across $\epsilon$ values. With SynTF and SanText, this phenomenon can be attributed to the mechanism design, as they operate differently from the five selected MLDP mechanisms. Furthermore, we see that the ratios of all methods generally follow the trend exhibited by the utility lines in Figure \ref{fig:acc_chart}, leading us to believe that utility and the $N_w:S_w$ ratio are closely correlated, as best exhibited by the Gumbel mechanism.

An important discussion comes with the clear relationship between a characterization of \say{effective} perturbation, where one may base the effectiveness of a mechanism in preserving privacy by the level of \textit{plausible deniability} that it provides. Thus, a mechanism that has a high probability of perturbing words (low $N_w$) and a high \textit{diversity} of output words (high $S_w$) would present the most attractive option. However, one can observe that the mechanisms with the lowest ratio scores, e.g., the Vickrey mechanism, in Figure \ref{fig:ns_ratio} also demonstrate lower utility scores in Figure \ref{fig:acc_chart}. Mechanisms that operate in more of a \say{balanced} fashion, e.g., the Gumbel mechanism, suffer less from utility drops. This illustrates an important finding regarding mechanism design, namely that plausible deniability and utility preservation must be considered in parallel. 

In a similar vein, the MLR analysis gives interesting insights into the connection between privacy and utility. If we assume that $N_w$ and $S_w$, under a certain $\epsilon$ constraint, supplemented with the empirical observation of \textit{PP}, can characterize a general MLDP mechanism well, then one can very well predict $CS$ given any mechanism. This provides a useful link to mechanism design and the (predicted) effect on the preservation of semantics, and ultimately, the effect on utility.

\begin{table*}[htbp]
\resizebox{0.99\linewidth}{!}{
\begin{tabular}{r|llll}
\toprule
\multicolumn{1}{r|}{\textbf{Original:}} & \multicolumn{3}{c}{\textbf{Sorry, gave it a 1, which is the rating I give to movies on which I walk out or fall asleep. In this case I fell asleep 10 minutes from the end, really, really bored and not caring at all about what happened next}} \\ \hline
\multicolumn{1}{r|}{\textbf{Preproc:}} & \multicolumn{3}{c}{\textbf{sorry gave rating give movie walk fall asleep case fell asleep minute end really really bored not caring happened next}} \\ \hline
\multicolumn{1}{c}{\textbf{Mechanism}} & \multicolumn{1}{c}{\textbf{\textit{d}}} & \multicolumn{1}{c}{\textbf{$\epsilon$}} & \multicolumn{1}{c}{Sentence} \\
\midrule
\midrule
CMP & 300 & 1 & gr gft expectable chakra grandparent gored magritte noo sniper breakfast meh substantive paternal verifiably viking flute erm striker muddies shane \\
CMP & 300 & 5 & relay neighborhood crime dubai hiroshima vampyre sandal kilcher estimate evers studiously mib cowgirl puzzle coldest stv danube berkley pulitzer del \\
CMP & 300 & 10 & cardiff interpretation efficiency hollywood movie stooped reacting sleep case export asleep minute dillon unable lot bored interfere trainor depressive gunter \\
SynTF & 300 & 1 & meritless devote rat ease\_up flick walkway fall departed character diminish asleep mo cease actually in\_truth blase not deal materialize future \\
SynTF & 300 & 5 & disconsolate afford rat devote movie base\_on\_balls downfall gone vitrine fell departed narrow terminal in\_truth truly tire non lovingness take\_place future \\
SynTF & 300 & 10 & bad cave\_in blackleg ease\_up movie paseo hang at\_peace font precipitate asleep mo stop truly actually bore not manage happen succeeding \\
TEM & 300 & 1 & peckenpahs urchin iñárritu lansburys streaming clout goosebump kissed welcomed maggies iwhippedi waswell occuped damme unbielevable calligraphy cameraman nula sharie british \\
TEM & 300 & 5 & sorry gave rating give movie walk fall asleep case fell asleep minute end really really bored not caring happened next \\
TEM & 300 & 10 & sorry gave rating give movie walk fall asleep case fell asleep minute end really really bored not caring happened next \\
Mahalanobis & 300 & 1 & perspicacious peace gf pellet gomer gargan raspberry kursk no prime wisconsin pickier reddin salvific designer clunkers incursion martyr hurd umm \\
Mahalanobis & 300 & 5 & sincere ha hrithik goalkeeper batman innes pole astral bellucci visa disfigured clan wale geometry faceoff simon sharia humperdink von faulty \\
Mahalanobis & 300 & 10 & sad summary rating age movie mum monarch asleep psychiatric fell asleep goalkeeper talker silver prototype improvising office caring thigpen declined \\
SanText & 300 & 1 & sorry gave rating give granddaughter walk fall asleep case fell asleep minute end really workhorse generate agreeing caring happened next \\
SanText & 300 & 5 & sorry gave rating give movie walk fall asleep hypothetically huggable selina memorably end crucially really starstruck insisting caring happened next \\
SanText & 300 & 10 & sorry gave rating give movie begun fall asleep concluded fell asleep minute end really really bored be caring happened next \\
Gumbel & 300 & 1 & embarrassed give indicating gave filmed anyway though awake reason dropped woke minute end definitely certainly tired though evera happen week \\
Gumbel & 300 & 5 & disappointed put rating giving movie walking fall awake example falling asleep equalizer ended really know bored although elderly happened expected \\
Gumbel & 300 & 10 & ashamed giving rating given starred walked coming sleep example slid fortunately came however obviously certainly bored be loving happening take \\
Vickrey & 300 & 1 & dah mayoral herein wachowski address ee corneau blazing ketchup observatory curled verdi thematic zen materialises ishwar wrestlemania nicholsons sonja interference \\
Vickrey & 300 & 5 & pepe pota eavesdrops hatching stunt yeop traumatizing takoma detained factly picher hitch light englund encyclopedia glanced calcium ditzy pasta chromosome \\
Vickrey & 300 & 10 & miserable assertion plunging invocation jerker sabre competing appetizer homicide dated suspended sanchez levy go consistency scene entertained flawed dreamt cbs \\
\bottomrule
\end{tabular}
}
\caption{Example text output on the IMDb dataset, for all evaluated mechanisms and $\varepsilon$ values on 300-dimensional GloVe embeddings. The original dataset text, as well as the preprocessed text, is given.}
\label{tab:ex}
\end{table*}

\paragraph{Privacy and Utility in the Same Light}
Key to this discussion is also an analysis of the composite quantification of privacy and utility, which was aided by our introduced metric: the PUC score. In Figure \ref{fig:puc_chart}, we see that the PUC score can vary quite significantly with the choice of $\alpha$. As an example, with an $\alpha$ of 0.75, one can clearly see that the Gumbel Mechanism presents an attractive choice of method, and this is supported by the observed accuracy score of Figure \ref{fig:acc_chart}. However, if we tune the parameter more towards privacy, this mechanism falls significantly from the top position. Indeed, looking to the privacy results of Gumbel, the observed scores are not on par with the other selected methods. Most notably, $S_w$ tends to be quite low. Similar analyses using the PUC score can be performed, with the goal of tailoring the benchmark interpretation to the privacy preferences of the user.

An immediate challenge with the quantification of the privacy-utility trade-off in a comparative manner traces back to the foundations of Metric Differential Privacy. With the generalization of DP to metrics, the comparability of the $\varepsilon$ parameter across mechanisms becomes more difficult, especially for those operating in different metric spaces. Although we evaluate our selected MLDP mechanisms on discrete $\varepsilon$ values, a more calibrated evaluation presents a concrete opportunity for improving the comparability of word-level MLDP mechanisms.

\paragraph{The Question of Metrics}
Our work highlights the need for a qualified and agreed upon set of metrics, which are necessary in order to evaluate word-level MLDP metrics on a uniform platform. In addition, this need for evaluation extends beyond the word-level to all DP NLP methods. The challenges this brings are numerous, rooted in the core challenge discussed above, i.e., the comparability of $\varepsilon$. In this work, we aim to begin the discussion with a base set of metrics, which provide the foundation for further metrics, as well as the opportunity to validate the efficacy of these metrics.

To start such validation, we critically view some of the privacy metrics proposed here. The privacy metric of \textit{LOW} presents an interesting point of inquiry, as this score varies quite significantly between mechanisms and experiment runs. No discernible interpretation, therefore, can be drawn; thus, an analysis of the usefulness of such a score is a topic of future investigation; therefore, further studies into the usefulness of this metric, as well as other lexical- or syntactic-based metrics, would be well-served. In addition, it becomes very important for the field of evaluating text privatization to agree upon a standard set of privacy metrics, something that currently does not exist. Such standardization would be paramount in unifying the validation and evaluation of privacy in NLP.  


Another dimension of evaluation and metrics not directly covered in this work is that of \textit{semantic coherence} and \textit{readability}. Although our CS and PP metrics capture to a degree the \say{closeness} of the perturbed text to the original, a closer look at the perturbed outputs (see Table \ref{tab:ex}) illustrates that there is still much room for improvement. Therefore, going forward in evaluating DP mechanisms, a greater emphasis should be placed on producing readable, coherent privatized outputs. This is also supported by the analysis of \citeauthor{mattern2022limits} regarding the question of optimal text privatization.

\section{Conclusion}
\label{sec:conclusion}
In this work, we conduct a comparative analysis of seven word-level Differential Privacy mechanisms, motivated by a lack of uniformity in the evaluation of word-level MLDP methods. We design a multi-dimensional experimental setup, which evaluates our chosen methods on two NLP tasks, with three $\epsilon$ parameters and three GloVe embedding dimensions, resulting in a total of 126 data points. To aid in the analysis, we employ a combination of utility and privacy metrics, as well as a novel composite score to quantify the interplay between the two. Finally, we include a discussion and analysis of the observed results, with the goal of providing a basis upon which future works on word-level MLDP evaluation for NLP can build. To this end, we provide a full replication repository, which can be found at \url{https://github.com/sjmeis/MLDP}.

As additional points of future work, we suggest (1) a more in-depth critique of the merits of word-level MLDP, (2) a focus on improving syntactic and semantic coherence in DP text perturbations (see Table \ref{tab:ex}), (3) the evaluation of MLDP mechanisms on a more diverse set of tasks and model architectures, and (4) further refinements on the metric-driven quantification and understanding of the privacy-utility trade-off in the NLP domain.

The results of our comparative analysis show that the classic argument of \say{higher privacy, lower utility, and vice versa} is considerably more complex, particularly in the realm of text. The evaluation of word-level MLDP methods is a start to tackling this question, and its implications provide the impetus to the continued study of privacy in NLP.

\newpage

\section{Acknowledgements}
The authors would like to thank Maulik Chevli for his valuable contributions to this work.

\section{Limitations and Ethics Statement}
While the aim of our work was to provide a uniform and fair evaluation for word-level MLDP methods, a clear limitation comes with the selection of our experiment parameters, or rather, those that were left out. Firstly, only pre-trained GloVe embeddings were utilized; other models such as Word2Vec or FastText were not included. In addition, the choice of $\epsilon$ can be perceived as limiting, especially the lack of uniformity in $\epsilon$ selection, and the resulting effects. This is especially crucial with MLDP, as the underlying distance metric for each mechanism affects the scale of noise added. Finally, the limitation of computational resources did not allow us to fine-tune algorithm-specific parameters; thus, we chose a single value for such parameters. 

Other limitations include our choice of privacy metrics. Empirical Privacy (degradation of adversarial performance) was not measured. Furthermore, our chosen metrics included previously unused metrics, such as \textit{LOW}. While these metrics were seen to be useful for illustrative purposes, validation of them as useful metrics would be an excellent point of future work.

On the note of metrics, our proposed PUC Score assumes that all individual metrics included in the composite scoring are weighted equally, an assumption that may or may not reflect the preferences or requirements of real-world practitioners. Thus, further work in the refinement of the composite metric to allow for such flexibility is needed.

As a benchmarking study, a final limitation comes with the fact that we did not benchmark time or resource consumption for each of our experimental runs. A quantification of the time needed, as well as the computational overhead, to perform word-level MLDP would be very useful in completing the picture we present in this work. 

Regarding ethical implications, the core of this study looks to an increasingly important topic of societal relevance, namely that of data privacy. In this light, we hope that our work contributes to the principle of respecting privacy in the processing of text data, particularly that which may contain sensitive information.

An ethical consideration to note is using pre-trained word embedding models as the basis for word-level MLDP. Possible ethical concerns with such models have been pointed out \cite{10.5555/3157382.3157584, 10.1145/3351095.3372843, manzini2019black}, and the effect of bias in these models was not tested by our study. 

\newpage
\nocite{*}
\section{Bibliographical References}\label{sec:reference}

\bibliographystyle{lrec-coling2024-natbib}
\bibliography{lrec-coling2024-example}


\appendix
\onecolumn

\section{MLDP Mechanism Pseudocode}
\label{sec:pseudocode}

\begin{algorithm}[htbp]
\caption*{\textbf{Algorithm:} SynTF Term-Frequency Vector Synthesis \cite{weggenmann2018syntf}}
\begin{algorithmic}[1]
\Require Document vector $\theta_t$, desired output length $n$, 
        privacy parameter $\epsilon > 0$, vocabulary $V$, rating function $\rho: V \times V \rightarrow [0,1]$.
\Ensure Synthetic tf vector $s \in \mathbb{N}^{|V|}$ with $|s| = n$
\For{$i \in \{1, \ldots, n\}$} 
\State Randomly sample word $v_i = R\ \text{Cat}(\theta_t)$ 
\State Randomly choose synonym $w_i = R\ \text{E}_\epsilon, \rho(v_i)$
\EndFor
\State $s_w = |{i \in [1, n]: w_i = w}|$ for all $w \in V$ \Comment{count synonyms}
\State \Return $s$
\end{algorithmic}
\end{algorithm}

\begin{algorithm}[htbp]
\caption*{\textbf{Algorithm:} Calibrated Multivariate Perturbation Mechanism \cite{feyisetan2020privacy}}
\begin{algorithmic}[1]
\Require String $x = w_1 w_2 \ldots w_n$, privacy parameter $\epsilon > 0$, word set $\mathcal{W}$.
\Ensure Privatized string $\tilde{s}$
\For{$i \in \{1, \ldots, n\}$}
\State Compute embedding $\varphi_i = \varphi(w_i)$ 
\State Perturb embedding to obtain $\hat{\varphi}_i = \varphi_i + \mathcal{N}$ with noise density $p_\mathcal{N}(z) \propto \exp(-\epsilon \Vert z\Vert)$ 
\State Obtain perturbed word $\hat{w}_i = \arg\min_{u\in \mathcal{W}} |\varphi(u) - \hat{\varphi}_i|$ 
\State Insert $\hat{w}_i$ in $i^{th}$ position of $\hat{x}$
\EndFor
\State \textbf{return} $\hat{x}$
\end{algorithmic}
\end{algorithm}

\begin{algorithm}[htbp]
\caption*{\textbf{Algorithm:} The Mahalanobis Mechanism \cite{xu2020differentially}}
\begin{algorithmic}[1]
\Require String $s = w_1w_2 \ldots w_n$, privacy parameter $\epsilon > 0$, scaled sample covariance matrix $\Sigma$, tuning parameter $\lambda \in [0, 1]$, word set $\mathcal{W}$.
\Ensure Privatized string $\tilde{s}$
\For{$i \in \{1, \ldots, n\}$}
\State Sample $Z$ from $f_Z(z) \propto \exp(-\epsilon \|z\|_{RM})$ 
\State Obtain the perturbed embedding $\hat{\phi}_i = \phi(w_i) + Z$.
\State Replace $w_i$ with $\hat{w}_i = \arg\min_{w\in \mathcal{W}} \|\phi(w) - \hat{\phi}_i\|_{2}$.
\EndFor
\State \textbf{return} $\tilde{s} = \hat{w}_1\hat{w}_2 \ldots \hat{w}_n$
\end{algorithmic}
\end{algorithm}

\begin{algorithm}[H]
\caption*{\textbf{Algorithm:} SanText Base Mechanism \cite{yue}}
\begin{algorithmic}[1]
    \Require A private document $D = \langle x_i \rangle_{i=1}^L$ and a privacy parameter $\epsilon > 0$.
    \Ensure Sanitized document $\hat{D}$
    \State Derive token vectors $\phi(x_i)$ for $i \in [1,L]$;
    \For{$i = 1,\dots,L$}
        \State Run $M(x_i)$ to sample sanitized token $y_i$ with $\Pr[M(x) = y] = C_x \cdot e^{-\frac{1}{2} \epsilon \cdot d_{\text{euc}}(\phi(x), \phi(y))}$
    \EndFor
    \State Output sanitized $\hat{D}$ as $\langle y_i \rangle_{i=1}^L$
\end{algorithmic}
\end{algorithm}

\begin{algorithm}[H]
\caption*{\textbf{Algorithm:} The Truncated Gumbel Mechanism \cite{xu3}}
\begin{algorithmic}[1]
\Require String $s = w_1w_2 \ldots w_n$, privacy parameter $\epsilon > 0$, word set $\mathcal{W}$.
\Ensure Privatized string $\tilde{s}$
\State $\Delta = \max_{w,w'\in \mathcal{W}}\Vert \phi(w) - \phi(w') \Vert$,  $\Delta_0 = \min_{w,w'\in \mathcal{W}}\Vert \phi(w) - \phi(w') \Vert$ \Comment{max, min word distance}
\State b = \(\frac{2\Delta}{\min\{W(2\alpha \Delta), \log_e(\alpha \Delta_0)\}}\), with $\alpha = \frac{1}{3}\Bigl(\epsilon - \frac{2(1+\log\Vert\mathcal{W}\Vert}{\Delta_0}\Bigr)$ and W = principal branch of Lambert W function.

\For{$i \in \{1, \ldots, n\}$}
\State $k \sim \mathrm{TruncatedPoisson}(\log|\mathcal{W}|;1,|\mathcal{W}|)$ and find $k$ closest words to $w_i$ as $\mathbf{u}$ with $u_1 = w_i$
\State Compute $k$ distances $\mathbf{d}$, where $d_j = \Vert w_i - u_j \Vert_2$
\State $\hat{w}_i = u_j$, where $j = \underset{\{d_1+g_1,...,d_k+g_k\}}{\argmin}$ and $g_1,...,g_k \sim_{i.i.d.} \mathrm{TruncatedGumbel}(0,b,\Delta)$
\EndFor
\State \textbf{return} $\tilde{s} = \hat{w}_1\hat{w}_2 \ldots \hat{w}_n$.
\end{algorithmic}
\end{algorithm}

\begin{algorithm}[htbp]
\caption*{\textbf{Algorithm:} The Vickrey Mechanism (k=2) \cite{xu2}}
\begin{algorithmic}[1]
\Require String $s = w_1w_2 \ldots w_n$, metric $d$, privacy parameter $\epsilon > 0$, tuning parameter $t \in [0,1]$, word set $\mathcal{W}$.
\Ensure Privatized string $\tilde{s}$
\For{$i \in \{1, \ldots, n\}$}
\State Sample $Z$ from $p(z) \propto \exp\{-\epsilon d(z,0)\}$
\State Obtain the perturbed embedding $\hat{\phi}_i = \phi(w_i) + Z$
\State Let $\tilde{w}_{i1} = \underset{w \in \mathcal{W}\setminus\{w_i\}}{\argmin}\Vert \hat{\phi}_i - \phi(w)\Vert_2$ and $\tilde{w}_{i2} = \underset{w \in \mathcal{W}\setminus\{w_i, \tilde{w}_{i1}\}}{\argmin}\Vert \hat{\phi}_i - \phi(w)\Vert_2$
\State
    $$\text{Set } \hat{w}_i = 
    \begin{cases}
        \tilde{w}_{i1}, & $\text{w/ Pr }$ p(t,\hat{\phi}_i) \\
        \tilde{w}_{i2}, & $\text{w/ Pr }$ 1 - p(t,\hat{\phi}_i)
    \end{cases},
    \text{where } p(t,\hat{\phi}_i) = \frac{(1-t)\Vert \phi(\tilde{w}_{i2})-\hat{\phi}_i\Vert_2}{t\Vert \phi(\tilde{w}_{i1})-\hat{\phi}_i\Vert_2 + (1-t)\Vert \phi(\tilde{w}_{i2})-\hat{\phi}_i\Vert_2}$$
\EndFor
\State \textbf{return} $\tilde{s} = \hat{w}_1\hat{w}_2 \ldots \hat{w}_n$.
\end{algorithmic}
\end{algorithm}

\begin{algorithm}[htbp]
\caption*{\textbf{Algorithm:} TEM: Metric Truncated Exponential Mechanism \cite{carvalho2023tem}}
\begin{algorithmic}[1]
\Require Word set $\mathcal{W}$, input word $w \in W$, truncation threshold $\gamma$, 
         metric $d_W: W \times W \rightarrow \mathbb{R}^+$, 
         and privacy parameter $\epsilon > 0$.
\Ensure Privatized string $\tilde{s}$
\State Given input $w$, obtain the set $L_w$ such that each word $w_i \in L_w$ satisfies $d_W(w, w_i) \leq \gamma$
\State Set the score $f(w,w_i)$ of each $w_i \in L_w$ as $f(w,w_i) = -d_W(w,w_i)$
\State Create a $\perp$ element with score $f(w,\perp) = -\gamma + 2\ln(|W \setminus L_w|)/\epsilon$
\For{each word $w_i \in L(x) \cup {\perp}$}
\State add Gumbel noise with mean $0$ and scale $2/\epsilon$ to score $f(w,w_i)$
\EndFor
\State Select $\hat{w}$ as the element with maximum noisy score from $L(x) \cup {\perp}$
\If{$\hat{w} = \perp$}
\State \textbf{return} random sample of $W \setminus L_w$
\Else
\State \textbf{return} $\hat{w}$
\EndIf
\end{algorithmic}
\end{algorithm}

\section{Privacy Metric Results}
\label{sec:priv_metrics}
In Tables \ref{tab:priv_imdb} and \ref{tab:priv_ag}, we present the complete set of privacy metrics as evaluated in our study. For each combination of (\textit{task, dimension, epsilon}), we \textbf{bold} the best score, which is governed by whether a lower ($\downarrow$) or a higher ($\uparrow$) score is better.
\begin{table}[htbp]
\resizebox{\linewidth}{!}{
\begin{tabular}{l|lllllllllllllll}
\toprule
\multicolumn{1}{r|}{\textbf{Task:}} & \multicolumn{15}{c}{\textbf{Sentiment Analysis (IMDb)}} \\ \hline
\multicolumn{1}{r|}{Metric:} & \multicolumn{1}{r}{PD ($N_w$) $\downarrow$} & \multicolumn{1}{r}{PD ($S_w$) $\uparrow$} & \multicolumn{1}{r}{PP $\uparrow$} & \multicolumn{1}{r}{CS $\uparrow$} & \multicolumn{1}{r|}{LOW $\downarrow$} & \multicolumn{1}{r}{PD ($N_w$) $\downarrow$} & \multicolumn{1}{r}{PD ($S_w$) $\uparrow$} & \multicolumn{1}{r}{PP $\uparrow$} & \multicolumn{1}{r}{CS $\uparrow$} & \multicolumn{1}{r|}{LOW $\downarrow$} & \multicolumn{1}{r}{PD ($N_w$) $\downarrow$} & \multicolumn{1}{r}{PD ($S_w$) $\uparrow$} & \multicolumn{1}{r}{PP $\uparrow$} & \multicolumn{1}{r}{CS $\uparrow$} & \multicolumn{1}{r}{LOW $\downarrow$} \\ \hline
\multicolumn{1}{r|}{Epsilon:} & \multicolumn{5}{c|}{1} & \multicolumn{5}{c|}{5} & \multicolumn{5}{c}{10} \\
\midrule
\midrule
\multicolumn{1}{r|}{Dimension:} & \multicolumn{15}{c}{50} \\ \hline
SynTF & 32.1 & 5.1 & 70.5 & 62.9 & \multicolumn{1}{l|}{70.5} & 33.0 & 5.1 & 76.5 & 61.6 & \multicolumn{1}{l|}{77.1} & 33.4 & 5.1 & 78.4 & 62.2 & 76.6 \\
CMP & \textbf{0.0} & 97.5 & \textbf{98.2} & 33.5 & \multicolumn{1}{l|}{46.8} & 9.1 & 87.8 & 90.1 & 45.0 & \multicolumn{1}{l|}{47.2} & 74.1 & 24.7 & 33.5 & 82.0 & 64.3 \\
Mahalanobis & \textbf{0.0} & 97.4 & \textbf{98.2} & 33.9 & \multicolumn{1}{l|}{46.8} & 9.0 & 88.1 & 91.8 & 44.0 & \multicolumn{1}{l|}{46.9} & 71.2 & 26.6 & 42.7 & 76.7 & 56.6 \\
SanText & 14.6 & 84.8 & 65.0 & \textbf{70.7} & \multicolumn{1}{l|}{65.0} & 14.6 & 84.8 & 59.8 & 79.5 & \multicolumn{1}{l|}{\textbf{13.3}} & 14.0 & \textbf{85.4} & 59.2 & 82.5 & \textbf{29.1} \\
Gumbel & 23.6 & 13.7 & 76.6 & 64.0 & \multicolumn{1}{l|}{53.1} & 22.7 & 13.3 & 76.8 & 63.9 & \multicolumn{1}{l|}{52.5} & 24.0 & 13.6 & 76.9 & 63.8 & 53.6 \\
Vickrey & \textbf{0.0} & 99.0 & \textbf{98.2} & 33.6 & \multicolumn{1}{l|}{46.7} & \textbf{3.5} & \textbf{91.8} & \textbf{95.0} & 42.8 & \multicolumn{1}{l|}{46.9} & \textbf{5.8} & 64.9 & \textbf{91.0} & 53.9 & 47.5 \\
TEM & \textbf{0.0} & \textbf{99.9} & \textbf{98.2} & 35.5 & \multicolumn{1}{l|}{\textbf{2.6}} & 68.1 & 31.8 & 18.4 & \textbf{87.3} & \multicolumn{1}{l|}{30.5} & 99.9 & 0.1 & 0.1 & \textbf{99.0} & 99.8 \\ \hline
\multicolumn{1}{r|}{Dimension:} & \multicolumn{15}{c}{100} \\ \hline
SynTF & 31.3 & 5.1 & 70.3 & 62.9 & \multicolumn{1}{l|}{77.0} & 32.1 & 5.1 & 76.2 & 61.8 & \multicolumn{1}{l|}{76.8} & 32.4 & 5.1 & 78.1 & 62.4 & 77.2 \\
CMP & \textbf{0.0} & 97.3 & \textbf{98.2} & 33.9 & \multicolumn{1}{l|}{46.9} & 3.4 & 95.0 & 94.4 & 40.6 & \multicolumn{1}{l|}{46.7} & 52.2 & 46.0 & 47.1 & 72.5 & 48.0 \\
Mahalanobis & \textbf{0.0} & 98.1 & \textbf{98.2} & 34.2 & \multicolumn{1}{l|}{46.7} & 3.3 & 95.8 & 95.2 & 39.8 & \multicolumn{1}{l|}{46.7} & 46.9 & 51.2 & 58.6 & 65.2 & 48.3 \\
SanText & 13.6 & 86.0 & 66.5 & \textbf{70.6} & \multicolumn{1}{l|}{\textbf{3.0}} & 14.4 & 85.2 & 59.4 & 79.4 & \multicolumn{1}{l|}{\textbf{9.9}} & 14.0 & \textbf{85.2} & 59.2 & 83.1 & \textbf{26.8} \\
Gumbel & 28.0 & 13.9 & 73.0 & 67.2 & \multicolumn{1}{l|}{42.6} & 27.2 & 13.8 & 73.4 & 67.0 & \multicolumn{1}{l|}{42.3} & 25.3 & 13.8 & 73.7 & 66.8 & 41.3 \\
Vickrey & \textbf{0.0} & 99.0 & \textbf{98.2} & 34.0 & \multicolumn{1}{l|}{46.8} & \textbf{2.0} & \textbf{96.7} & \textbf{96.5} & 39.4 & \multicolumn{1}{l|}{47.1} & \textbf{6.4} & 80.2 & \textbf{90.5} & 51.2 & 46.9 \\
TEM & \textbf{0.0} & \textbf{99.9} & \textbf{98.2} & 35.5 & \multicolumn{1}{l|}{3.1} & 77.5 & 22.4 & 6.1 & \textbf{95.1} & \multicolumn{1}{l|}{37.5} & 99.9 & 0.1 & 0.1 & \textbf{99.0} & 97.9 \\ \hline
\multicolumn{1}{r|}{Dimension:} & \multicolumn{15}{c}{300} \\ \hline
SynTF & 32.1 & 5.0 & 69.9 & 63.2 & \multicolumn{1}{l|}{76.7} & 32.0 & 5.1 & 75.4 & 62.1 & \multicolumn{1}{l|}{76.3} & 32.6 & 5.1 & 77.5 & 62.8 & 76.8 \\
CMP & \textbf{0.0} & 98.0 & \textbf{98.2} & 32.7 & \multicolumn{1}{l|}{46.9} & 1.0 & 98.7 & 97.4 & 35.6 & \multicolumn{1}{l|}{46.8} & 27.8 & 71.5 & 78.8 & 49.8 & 46.7 \\
Mahalanobis & \textbf{0.0} & 98.5 & \textbf{98.2} & 34.0 & \multicolumn{1}{l|}{46.9} & 0.6 & 99.1 & \textbf{97.7} & 36.2 & \multicolumn{1}{l|}{46.8} & 20.4 & 78.9 & 88.5 & 44.2 & 46.5 \\
SanText & 14.4 & 85.3 & 66.6 & \textbf{70.6} & \multicolumn{1}{l|}{\textbf{2.7}} & 14.2 & 85.3 & 59.2 & 80.7 & \multicolumn{1}{l|}{\textbf{16.6}} & 14.3 & 85.5 & 59.2 & 84.0 & 33.1 \\
Gumbel & 31.7 & 13.4 & 69.5 & 68.1 & \multicolumn{1}{l|}{25.2} & 32.5 & 13.0 & 70.0 & 67.8 & \multicolumn{1}{l|}{25.8} & 27.7 & 13.6 & 70.9 & 67.5 & \textbf{25.0} \\
Vickrey & \textbf{0.0} & 99.5 & \textbf{98.2} & 34.3 & \multicolumn{1}{l|}{48.3} & \textbf{0.5} & \textbf{99.2} & 97.63 & 37.2 & \multicolumn{1}{l|}{47.6} & \textbf{5.8} & \textbf{91.0} & \textbf{92.7} & 44.5 & 49.3 \\
TEM & 0.1 & \textbf{99.8} & 98.1 & 35.2 & \multicolumn{1}{l|}{3.0} & 99.2 & 0.8 & 1.4 & \textbf{98.7} & \multicolumn{1}{l|}{71.4} & 99.9 & 0.1 & 0.1 & \textbf{99.9} & 98.4 \\ \bottomrule
\end{tabular}
}
\caption{Privacy scores for the Sentiment Anaylsis (IMDb) experiments.}
\label{tab:priv_imdb}
\end{table}

\begin{table}[htbp]
\resizebox{\linewidth}{!}{
\begin{tabular}{l|lllllllllllllll}
\toprule
\multicolumn{1}{r|}{\textbf{Task:}} & \multicolumn{15}{c}{\textbf{Topic Classification (AG News)}} \\ \hline
\multicolumn{1}{r|}{Metric:} & \multicolumn{1}{r}{PD ($N_w$) $\downarrow$} & \multicolumn{1}{r}{PD ($S_w$) $\uparrow$} & \multicolumn{1}{r}{PP $\uparrow$} & \multicolumn{1}{r}{CS $\uparrow$} & \multicolumn{1}{r|}{LOW $\downarrow$} & \multicolumn{1}{r}{PD ($N_w$) $\downarrow$} & \multicolumn{1}{r}{PD ($S_w$) $\uparrow$} & \multicolumn{1}{r}{PP $\uparrow$} & \multicolumn{1}{r}{CS $\uparrow$} & \multicolumn{1}{r|}{LOW $\downarrow$} & \multicolumn{1}{r}{PD ($N_w$) $\downarrow$} & \multicolumn{1}{r}{PD ($S_w$) $\uparrow$} & \multicolumn{1}{r}{PP $\uparrow$} & \multicolumn{1}{r}{CS $\uparrow$} & \multicolumn{1}{r}{LOW $\downarrow$} \\ \hline
\multicolumn{1}{r|}{Epsilon:} & \multicolumn{5}{c|}{1} & \multicolumn{5}{c|}{5} & \multicolumn{5}{c}{10} \\
\midrule
\midrule
\multicolumn{1}{r|}{Dimension:} & \multicolumn{15}{c}{50} \\ \hline
SynTF & 55.1 & 4.6 & 60.7 & \textbf{61.3} & \multicolumn{1}{l|}{61.7} & 54.8 & 4.7 & 66.4 & 58.6 & \multicolumn{1}{l|}{60.3} & 55.2 & 4.8 & 68.4 & 58.1 & 60.2 \\
CMP & \textbf{0.0} & 97.4 & \textbf{98.8} & 20.3 & \multicolumn{1}{l|}{16.0} & 12.4 & 83.9 & 85.9 & 37.9 & \multicolumn{1}{l|}{17.2} & 75.8 & 21.3 & 23.1 & 84.5 & 52.5 \\
Mahalanobis & \textbf{0.0} & 97.7 & \textbf{98.8} & 20.9 & \multicolumn{1}{l|}{16.5} & 10.9 & 85.3 & 88.4 & 35.5 & \multicolumn{1}{l|}{16.2} & 75.0 & 22.2 & 30.3 & 79.3 & 36.8 \\
SanText & 3.9 & 95.0 & 47.2 & 60.0 & \multicolumn{1}{l|}{5.7} & \textbf{3.2} & \textbf{95.7} & 26.5 & 81.0 & \multicolumn{1}{l|}{27.5} & \textbf{3.0} & \textbf{95.0} & 25.7 & 84.3 & 25.7 \\
Gumbel & 25.6 & 13.2 & 73.0 & 57.7 & \multicolumn{1}{l|}{27.2} & 26.9 & 13.0 & 73.2 & 57.6 & \multicolumn{1}{l|}{29.3} & 26.0 & 12.7 & 73.4 & 57.6 & 28.2 \\
Vickrey & \textbf{0.0} & 98.8 & \textbf{98.8} & 20.8 & \multicolumn{1}{l|}{16.1} & 4.7 & 90.4 & \textbf{94.2} & 33.2 & \multicolumn{1}{l|}{\textbf{16.0}} & 5.1 & 57.6 & \textbf{92.8} & 44.0 & \textbf{20.0} \\
TEM & 0.1 & \textbf{99.7} & \textbf{98.8} & 21.2 & \multicolumn{1}{l|}{\textbf{4.5}} & 87.2 & 12.6 & 7.5 & \textbf{94.3} & \multicolumn{1}{l|}{69.3} & 100.0 & 0.0 & 0.5 & \textbf{99.9} & 99.3 \\ \hline
\multicolumn{1}{r|}{Dimension:} & \multicolumn{15}{c}{100} \\ \hline
SynTF & 55.4 & 4.5 & 60.6 & \textbf{61.4} & \multicolumn{1}{l|}{61.6} & 55.2 & 4.6 & 66.1 & 58.8 & \multicolumn{1}{l|}{59.8} & 55.4 & 4.6 & 68.1 & 58.3 & 59.9 \\
CMP & \textbf{0.0} & 97.9 & 98.8 & 21.1 & \multicolumn{1}{l|}{17.3} & 5.7 & 93.0 & 92.3 & 31.2 & \multicolumn{1}{l|}{16.7} & 60.2 & 37.2 & 35.3 & 75.0 & 25.1 \\
Mahalanobis & \textbf{0.0} & 98.3 & 98.8 & 21.5 & \multicolumn{1}{l|}{17.7} & 5.5 & 93.0 & 93.7 & 29.6 & \multicolumn{1}{l|}{\textbf{15.9}} & 58.8 & 38.9 & 46.5 & 66.4 & \textbf{17.7} \\
SanText & 3.9 & 95.4 & 47.2 & 59.8 & \multicolumn{1}{l|}{5.4} & 4.2 & 94.8 & 25.8 & 81.6 & \multicolumn{1}{l|}{20.8} & \textbf{4.7} & \textbf{94.6} & 25.6 & 85.0 & 56.4 \\
Gumbel & 29.8 & 13.0 & 70.7 & 60.8 & \multicolumn{1}{l|}{22.8} & 28.7 & 12.5 & 71.0 & 60.6 & \multicolumn{1}{l|}{21.5} & 27.6 & 12.6 & 71.5 & 60.3 & 22.3 \\
Vickrey & \textbf{0.0} & 98.9 & \textbf{98.9} & 21.3 & \multicolumn{1}{l|}{16.6} & \textbf{2.8} & \textbf{95.8} & \textbf{96.2} & 28.8 & \multicolumn{1}{l|}{16.9} & 6.8 & 78.1 & \textbf{91.7} & 41.5 & \textbf{17.7} \\
TEM & 0.1 & \textbf{99.7} & 98.8 & 20.8 & \multicolumn{1}{l|}{\textbf{4.4}} & 93.6 & 6.4 & 2.1 & \textbf{98.5} & \multicolumn{1}{l|}{81.4} & 99.8 & 0.2 & 0.2 & \textbf{99.9} & 98.3 \\ \hline
\multicolumn{1}{r|}{Dimension:} & \multicolumn{15}{c}{300} \\ \hline
SynTF & 55.7 & 4.6 & 60.2 & \textbf{61.6} & \multicolumn{1}{l|}{61.2} & 55.7 & 4.6 & 65.3 & 59.1 & \multicolumn{1}{l|}{61.7} & 55.4 & 4.7 & 67.7 & 58.5 & 60.2 \\
CMP & \textbf{0.0} & 98.1 & \textbf{98.9} & 19.1 & \multicolumn{1}{l|}{16.5} & 2.2 & 97.0 & 97.3 & 23.8 & \multicolumn{1}{l|}{16.0} & 34.9 & 63.7 & 69.2 & 48.3 & 15.6 \\
Mahalanobis & \textbf{0.0} & 98.2 & \textbf{98.9} & 20.8 & \multicolumn{1}{l|}{17.5} & 1.5 & 98.1 & 97.9 & 23.9 & \multicolumn{1}{l|}{16.4} & 26.8 & 71.8 & 83.5 & 36.9 & 15.0 \\
SanText & 3.6 & 95.5 & 47.2 & 59.8 & \multicolumn{1}{l|}{6.3} & 4.0 & 95.5 & 25.8 & 82.0 & \multicolumn{1}{l|}{32.8} & \textbf{3.6} & \textbf{95.7} & 25.7 & 85.4 & 57.7 \\
Gumbel & 32.0 & 12.6 & 68.8 & 60.5 & \multicolumn{1}{l|}{23.9} & 31.2 & 12.8 & 69.2 & 60.2 & \multicolumn{1}{l|}{23.5} & 30.7 & 13.2 & 69.9 & 59.6 & 23.9 \\
Vickrey & \textbf{0.0} & 98.8 & \textbf{98.9} & 19.6 & \multicolumn{1}{l|}{16.8} & \textbf{1.0} & \textbf{98.5} & \textbf{98.0} & 23.5 & \multicolumn{1}{l|}{\textbf{15.3}} & 6.6 & 89.7 & \textbf{92.5} & 33.6 & \textbf{14.6} \\
TEM & 0.4 & \textbf{99.4} & 98.6 & 20.5 & \multicolumn{1}{l|}{\textbf{5.4}} & 99.9 & 0.1 & 0.51 & \textbf{99.8} & \multicolumn{1}{l|}{98.0} & 99.8 & 0.2 & 0.13 & \textbf{99.9} & 98.4 \\ 
\bottomrule
\end{tabular}
}
\caption{Privacy scores for the Topic Classification (AG News) experiments.}
\label{tab:priv_ag}
\end{table}

\subsection{Percentage of English Words}
As another metric to measure the effect of MLDP methods on word perturbation, we measured the percentage of English words existing in the perturbed datasets, as opposed to the baseline in the original ones. To calculate this metric, we leveraged the \textbf{words} corpus from \textit{NLTK}. Note that a word not being counted as English does not necessarily mean that it is in a different language, but rather that it is not included as a standard word in our chosen corpus of English words.

As the true relation between privacy preservation via word perturbations and the resulting effect on English words is not well known and justifiable in comparison to our other metrics, we exclude this metric from our main analysis and instead include the results here. Nevertheless, it is interesting to observe that some methods significantly reduce the number of English words (e.g., CMP and TEM). A more in-depth study into the relevance of these results can be a point of future investigation.

\begin{table}[htbp]
\centering
\begin{tabular}{l|lll||lll}
\toprule
\multicolumn{1}{r|}{Dataset} & \multicolumn{3}{c||}{IMDb} & \multicolumn{3}{c}{AG News} \\ \hline
\multicolumn{1}{r|}{Baseline} & \multicolumn{3}{c||}{84.75} & \multicolumn{3}{c}{73.63} \\ \hline
\multicolumn{1}{r|}{Epsilon} & \multicolumn{1}{c}{1} & \multicolumn{1}{c}{5} & \multicolumn{1}{c||}{10} & \multicolumn{1}{c}{1} & \multicolumn{1}{c}{5} & \multicolumn{1}{c}{10} \\ 
\midrule
SynTF & \multicolumn{1}{c}{85.05} & \multicolumn{1}{c}{85.92} & \multicolumn{1}{c||}{86.46} & \multicolumn{1}{c}{75.16} & \multicolumn{1}{c}{75.62} & \multicolumn{1}{c}{75.96} \\ 
CMP & \multicolumn{1}{c}{53.97} & \multicolumn{1}{c}{67.77} & \multicolumn{1}{c||}{78.08} & \multicolumn{1}{c}{41.55} & \multicolumn{1}{c}{59.88} & \multicolumn{1}{c}{69.74} \\ 
Mahalanobis & \multicolumn{1}{c}{57.44} & \multicolumn{1}{c}{67.67} & \multicolumn{1}{c||}{75.72} & \multicolumn{1}{c}{47.67} & \multicolumn{1}{c}{60.13} & \multicolumn{1}{c}{67.54} \\ 
SanText & \multicolumn{1}{c}{76.42} & \multicolumn{1}{c}{78.33} & \multicolumn{1}{c||}{80.54} & \multicolumn{1}{c}{67.36} & \multicolumn{1}{c}{70.98} & \multicolumn{1}{c}{71.40} \\ 
Gumbel & \multicolumn{1}{c}{79.18} & \multicolumn{1}{c}{79.13} & \multicolumn{1}{c||}{79.10} & \multicolumn{1}{c}{67.18} & \multicolumn{1}{c}{67.16} & \multicolumn{1}{c}{67.05} \\ 
Vickrey & \multicolumn{1}{c}{56.95} & \multicolumn{1}{c}{69.00} & \multicolumn{1}{c||}{75.76} & \multicolumn{1}{c}{44.40} & \multicolumn{1}{c}{59.97} & \multicolumn{1}{c}{66.22} \\ 
TEM & \multicolumn{1}{c}{34.59} & \multicolumn{1}{c}{80.90} & \multicolumn{1}{c||}{84.73} & \multicolumn{1}{c}{45.24} & \multicolumn{1}{c}{72.92} & \multicolumn{1}{c}{73.60} \\ 
\bottomrule
\end{tabular}
\caption{Percentage of English words in perturbed datasets.}
\label{tab:english}
\end{table}

\subsection{Privacy-Utility Composite Scores}
The full set of Privacy-Utility Composite (PUC) scores is provided in Table \ref{tab:puc}.

\begin{table}[htbp]
\centering
\resizebox{0.95\linewidth}{!}{
\begin{tabular}{llllllllll||lllllllll}
\toprule
\multicolumn{1}{r|}{Task:} & \multicolumn{9}{c||}{\textbf{Sentiment Analysis (IMDb)}} & \multicolumn{9}{c}{\textbf{Topic Classification (AG News)}} \\ \hline
\multicolumn{1}{r|}{Tuning parameter:} & \multicolumn{3}{c|}{$\alpha=0.75$} & \multicolumn{3}{c|}{$\alpha=0.5$} & \multicolumn{3}{c||}{$\alpha=0.25$} & \multicolumn{3}{c|}{$\alpha=0.75$} & \multicolumn{3}{c|}{$\alpha=0.5$} & \multicolumn{3}{c}{$\alpha=0.25$} \\ \hline
\multicolumn{1}{r|}{Epsilon:} & \multicolumn{1}{c}{1} & \multicolumn{1}{c}{5} & \multicolumn{1}{c|}{10} & \multicolumn{1}{c}{1} & \multicolumn{1}{c}{5} & \multicolumn{1}{c|}{10} & \multicolumn{1}{c}{1} & \multicolumn{1}{c}{5} & \multicolumn{1}{c||}{10} & \multicolumn{1}{c}{1} & \multicolumn{1}{c}{5} & \multicolumn{1}{c|}{10} & \multicolumn{1}{c}{1} & \multicolumn{1}{c}{5} & \multicolumn{1}{c|}{10} & \multicolumn{1}{c}{1} & \multicolumn{1}{c}{5} & \multicolumn{1}{c}{10} \\ 
\midrule
\midrule
\multicolumn{1}{r|}{Dimension:} & \multicolumn{9}{c||}{50} & \multicolumn{9}{c}{50} \\ \hline
\multicolumn{1}{l|}{SynTF} & 82.22 & 81.76 & \multicolumn{1}{l|}{81.48} & 70.54 & 70.05 & \multicolumn{1}{l|}{70.03} & 58.86 & 58.33 & 58.59 & 78.98 & 92.25 & \multicolumn{1}{l|}{\textbf{93.01}} & \textbf{80.33} & \textbf{87.97} & \multicolumn{1}{l|}{\textbf{86.63}} & \textbf{81.67} & \textbf{83.70} & \textbf{80.24} \\
\multicolumn{1}{l|}{CMP} & 69.67 & 78.00 & \multicolumn{1}{l|}{85.19} & 71.94 & 76.44 & \multicolumn{1}{l|}{70.24} & 74.21 & 74.88 & 55.30 & \textbf{89.78} & \textbf{93.77} & \multicolumn{1}{l|}{90.42} & 79.26 & 87.72 & \multicolumn{1}{l|}{79.58} & 68.74 & 81.67 & 68.74 \\
\multicolumn{1}{l|}{Mahalanobis} & 70.75 & 83.12 & \multicolumn{1}{l|}{83.92} & 72.68 & 79.94 & \multicolumn{1}{l|}{70.49} & 74.61 & 76.77 & 57.07 & 77.86 & 90.55 & \multicolumn{1}{l|}{74.64} & 78.08 & 85.84 & \multicolumn{1}{l|}{74.84} & 78.3 & 81.13 & 75.05 \\
\multicolumn{1}{l|}{SanText} & 74.26 & 72.79 & \multicolumn{1}{l|}{72.10} & 72.23 & 74.94 & \multicolumn{1}{l|}{73.67} & 70.21 & 77.09 & 75.23 & 76.93 & 89.41 & \multicolumn{1}{l|}{86.36} & 77.98 & 78.78 & \multicolumn{1}{l|}{72.24} & 79.04 & 68.15 & 58.12 \\
\multicolumn{1}{l|}{Gumbel} & \textbf{89.64} & \textbf{88.60} & \multicolumn{1}{l|}{88.86} & \textbf{78.26} & 77.65 & \multicolumn{1}{l|}{77.69} & 66.89 & 66.71 & 66.51 & 76.68 & 75.74 & \multicolumn{1}{l|}{85.64} & 77.94 & 75.33 & \multicolumn{1}{l|}{70.47} & 79.20 & 74.91 & 55.29 \\
\multicolumn{1}{l|}{Vickrey} & 71.77 & 86.88 & \multicolumn{1}{l|}{\textbf{90.75}} & 73.46 & \textbf{83.20} & \multicolumn{1}{l|}{\textbf{84.27}} & 75.14 & \textbf{79.52} & \textbf{77.78} & 76.05 & 83.61 & \multicolumn{1}{l|}{83.23} & 77.43 & 70.05 & \multicolumn{1}{l|}{69.88} & 78.8 & 56.48 & 56.53 \\
\multicolumn{1}{l|}{TEM} & 72.37 & 84.23 & \multicolumn{1}{l|}{79.00} & 76.98 & 72.08 & \multicolumn{1}{l|}{59.3} & \textbf{81.59} & 59.93 & 39.60 & 83.53 & 83.59 & \multicolumn{1}{l|}{81.11} & 69.68 & 66.25 & \multicolumn{1}{l|}{60.81} & 55.82 & 48.92 & 40.52 \\ \hline
\multicolumn{1}{r|}{Dimension:} & \multicolumn{9}{c|}{100} & \multicolumn{9}{c}{100} \\ \hline
\multicolumn{1}{l|}{SynTF} & 81.19 & 81.99 & \multicolumn{1}{l|}{81.94} & 69.46 & 70.28 & \multicolumn{1}{l|}{70.36} & 57.73 & 58.56 & 77.00 & 80.24 & \textbf{91.07} & \multicolumn{1}{l|}{\textbf{93.92}} & \textbf{79.70} & \textbf{86.99} & \multicolumn{1}{l|}{\textbf{88.40}} & 79.16 & 82.90 & \textbf{82.88} \\
\multicolumn{1}{l|}{CMP} & 65.74 & 78.34 & \multicolumn{1}{l|}{87.37} & 69.32 & 77.55 & \multicolumn{1}{l|}{75.94} & 72.91 & 76.77 & 64.51 & \textbf{90.33} & 90.29 & \multicolumn{1}{l|}{89.33} & 79.68 & 86.52 & \multicolumn{1}{l|}{79.19} & 69.03 & 82.75 & 69.04 \\
\multicolumn{1}{l|}{Mahalanobis} & 65.99 & 74.44 & \multicolumn{1}{l|}{85.64} & 69.58 & 75.01 & \multicolumn{1}{l|}{75.75} & 73.17 & 75.59 & 65.85 & 76.18 & 88.40 & \multicolumn{1}{l|}{88.11} & 78.44 & 85.67 & \multicolumn{1}{l|}{77.09} & \textbf{80.70} & \textbf{82.95} & 66.08 \\
\multicolumn{1}{l|}{SanText} & 80.87 & 79.34 & \multicolumn{1}{l|}{77.12} & \textbf{80.95} & 79.54 & \multicolumn{1}{l|}{77.19} & 81.02 & \textbf{79.74} & 77.27 & 75.78 & 89.91 & \multicolumn{1}{l|}{87.95} & 77.25 & 79.53 & \multicolumn{1}{l|}{76.11} & 78.71 & 69.16 & 64.28 \\
\multicolumn{1}{l|}{Gumbel} & \textbf{87.88} & \textbf{88.57} & \multicolumn{1}{l|}{\textbf{90.51}} & 77.49 & 78.03 & \multicolumn{1}{l|}{79.52} & 67.09 & 67.48 & 68.53 & 73.22 & 80.7 & \multicolumn{1}{l|}{78.47} & 75.65 & 78.94 & \multicolumn{1}{l|}{75.26} & 78.07 & 77.19 & 72.04 \\
\multicolumn{1}{l|}{Vickrey} & 66.68 & 84.45 & \multicolumn{1}{l|}{88.29} & 70.08 & \textbf{81.87} & \multicolumn{1}{l|}{\textbf{83.43}} & 73.48 & 79.28 & \textbf{78.58} & 72.16 & 84.03 & \multicolumn{1}{l|}{83.36} & 74.81 & 70.32 & \multicolumn{1}{l|}{69.95} & 77.45 & 56.61 & 56.55 \\
\multicolumn{1}{l|}{TEM} & 73.35 & 83.26 & \multicolumn{1}{l|}{79.07} & 77.60 & 69.41 & \multicolumn{1}{l|}{59.48} & \textbf{81.85} & 55.57 & 39.88 & 85.4 & 81.43 & \multicolumn{1}{l|}{80.79} & 70.90 & 63.09 & \multicolumn{1}{l|}{60.67} & 56.4 & 44.74 & 40.56 \\ \hline
\multicolumn{1}{r}{Dimension:} & \multicolumn{9}{c|}{300} & \multicolumn{9}{c}{300} \\ \hline
\multicolumn{1}{l|}{SynTF} & 78.86 & 79.14 & \multicolumn{1}{l|}{79.44} & 67.86 & 68.38 & \multicolumn{1}{l|}{68.70} & 56.86 & 57.62 & 57.95 & 84.77 & 82.21 & \multicolumn{1}{l|}{92.65} & \textbf{82.69} & \textbf{81.72} & \multicolumn{1}{l|}{\textbf{88.07}} & \textbf{80.60} & \textbf{81.23} & \textbf{83.50} \\
\multicolumn{1}{l|}{CMP} & 64.22 & 69.59 & \multicolumn{1}{l|}{82.82} & 68.28 & 71.99 & \multicolumn{1}{l|}{76.92} & 72.34 & 74.38 & 71.02 & \textbf{88.56} & 81.89 & \multicolumn{1}{l|}{\textbf{93.83}} & 78.11 & 81.25 & \multicolumn{1}{l|}{85.91} & 67.65 & 80.62 & 78.00 \\
\multicolumn{1}{l|}{Mahalanobis} & 65.33 & 67.24 & \multicolumn{1}{l|}{77.48} & 69.14 & 70.53 & \multicolumn{1}{l|}{74.63} & 72.95 & 73.83 & 71.79 & 74.14 & 80.90 & \multicolumn{1}{l|}{89.91} & 76.94 & 80.73 & \multicolumn{1}{l|}{81.98} & 79.74 & 80.57 & 74.06 \\
\multicolumn{1}{l|}{SanText} & 83.45 & 80.89 & \multicolumn{1}{l|}{80.62} & \textbf{82.66} & 80.22 & \multicolumn{1}{l|}{\textbf{79.17}} & \textbf{81.87} & \textbf{79.55} & \textbf{77.71} & 67.23 & 84.25 & \multicolumn{1}{l|}{89.77} & 71.51 & 80.60 & \multicolumn{1}{l|}{79.05} & 75.80 & 76.95 & 68.34 \\
\multicolumn{1}{l|}{Gumbel} & \textbf{87.14} & \textbf{86.56} & \multicolumn{1}{l|}{\textbf{87.05}} & 77.70 & \textbf{77.21} & \multicolumn{1}{l|}{77.98} & 68.26 & 67.85 & 68.92 & 84.95 & \textbf{89.84} & \multicolumn{1}{l|}{83.22} & 70.6 & 79.06 & \multicolumn{1}{l|}{78.52} & 56.25 & 68.28 & 73.81 \\
\multicolumn{1}{l|}{Vickrey} & 66.80 & 71.64 & \multicolumn{1}{l|}{80.08} & 70.11 & 73.49 & \multicolumn{1}{l|}{78.26} & 73.43 & 75.33 & 76.44 & 65.63 & 84.08 & \multicolumn{1}{l|}{84.08} & 70.39 & 70.16 & \multicolumn{1}{l|}{70.41} & 75.15 & 56.24 & 56.73 \\
\multicolumn{1}{l|}{TEM} & 66.16 & 79.18 & \multicolumn{1}{l|}{78.91} & 72.78 & 61.47 & \multicolumn{1}{l|}{59.39} & 79.39 & 43.77 & 39.88 & 61.22 & 79.88 & \multicolumn{1}{l|}{77.85} & 67.51 & 60.08 & \multicolumn{1}{l|}{58.7} & 73.81 & 40.29 & 39.55 \\
\bottomrule
\end{tabular}
}
\caption{Privacy-Utility Composite (PUC) scores.}
\label{tab:puc}
\end{table}

\newpage
\section{Perturbation Examples}
\label{sec:examples}

Here, we display two representative samples of perturbed text under differing parameters and mechanisms, both for an IMDb  (Table \ref{tab:ex1}) and AG News (Table \ref{tab:ex2}) sentence.

\begin{table}[htbp]
\resizebox{0.99\linewidth}{!}{
\begin{tabular}{r|llll}
\toprule
\multicolumn{1}{r|}{\textbf{Original:}} & \multicolumn{3}{c}{\textbf{Sorry, gave it a 1, which is the rating I give to movies on which I walk out or fall asleep. In this case I fell asleep 10 minutes from the end, really, really bored and not caring at all about what happened next}} \\ \hline
\multicolumn{1}{r|}{\textbf{Preproc:}} & \multicolumn{3}{c}{\textbf{sorry gave rating give movie walk fall asleep case fell asleep minute end really really bored not caring happened next}} \\ \hline
\multicolumn{1}{c}{\textbf{Mechanism}} & \multicolumn{1}{c}{\textbf{\textit{d}}} & \multicolumn{1}{c}{\textbf{$\epsilon$}} & \multicolumn{1}{c}{Sentence} \\
\midrule
\midrule
CMP & 50 & 1 & miniseries sphincter malaysia victory ambassador worldfest deuce hah humid parliament uranium sergei inextricably cleric genus belleau bigalow deck wistfully expectable \\
CMP & 50 & 5 & punish relay brunt ready channel sky fall valve face vantage recordist flicked stretch e feel iago serious family happened ford \\
CMP & 50 & 10 & luckily gave rating extra movie halfway fall asleep case unchanged awake minute closing really good bored unable caring happened opening \\
SynTF & 50 & 1 & dingy chip\_in military\_rank cave\_in motion\_picture pass devolve gone encase return gone minute\_of\_arc ending really genuinely drill non give\_care befall adjacent \\
SynTF & 50 & 5 & deplorable have grade cave\_in flick walk\_of\_life descend at\_peace vitrine return asleep minute terminate real genuinely drill non handle chance adjacent \\
SynTF & 50 & 10 & no-count sacrifice valuation grant flick walking come\_down departed pillowcase felled\_seam departed second cease truly rattling drill not handle occur following \\
TEM & 50 & 1 & opener anxiety debilitating ingenue diminutiveaggressive lewisburns forjust macha said expressionbr lebanese chipped connors crazed marriedthis atrocitybut personification staying yuunagi geology \\
TEM & 50 & 5 & sorry gave rating give movie walk fall asleep case fell asleep minute end really answer bored determined caring happened next \\
TEM & 50 & 10 & sorry gave rating give movie walk fall asleep case fell asleep minute end really really bored not caring happened next \\
Mahalanobis & 50 & 1 & meh fatty expectable stalag biggest hideout iscariot falkland consecutive brier baseman loch mumbai seeded ahmed mythology verifiably socorro eking emission \\
Mahalanobis & 50 & 5 & katrina directorial interception copy movie dedicated fashion knife related lower audible minute attendance feeling love wince always teach heimlich avenue \\
Mahalanobis & 50 & 10 & sorry protector rating give movie fiance craze asleep charged rose asleep minute forced getting really depressed necessarily caring happened next \\
SanText & 50 & 1 & predictably gave rating give movie deemed fall asleep grass fell asleep minute end devastatingly fury relating not caring happened znaimer \\
SanText & 50 & 5 & sorry gave expands give movie walk fall asleep case tumbled panicking scoring end so really bored not embracing feverishly next \\
SanText & 50 & 10 & sorry referring rating give movie approaching coincide asleep case fell woke minute end really so dreamer not nurturing happened next \\
Gumbel & 50 & 1 & awful picked rated need comedy walking coming waited prosecution soared woke half end always something boring could trusting happened set \\
Gumbel & 50 & 5 & pity took rating giving movie walking year slept case dipped suddenly superb coming sure really boring cannot caring exactly starting \\
Gumbel & 50 & 10 & sorry gave notch needed drama walk rise waited complaint dropped waking goal end thing lot scared nothing caring knew next \\
Vickrey & 50 & 1 & brokedown fwd satellite matondkar gardening spokesman blubber ej eruption readin hearkens uschi connecting schamus overly dentistry meh andalou deathly witching \\
Vickrey & 50 & 5 & heartache proposition score harry undertaking nap period ingrate rafter nice breech zenon sustained pretty sure suicidal earth grateful sinking first \\
Vickrey & 50 & 10 & stupid sunday downgrade without show cruising trend momentarily case tumbled awakens equalizer turning think think harder say physically happen scheduled \\
CMP & 100 & 1 & undocumented puree bernarda labor blanka buh towel gf hah roof million larceny cholera kool duster shawn homeless ankle helfgott jazz \\
CMP & 100 & 5 & simba attract haunted tech gaby theresa pork scary charity testimony down save nile exerting percent laced splicing simmering offer syria \\
CMP & 100 & 10 & sorry appeared rating trying movie fancy distort climb case northern asleep minute end probably pretend furious publicly caring surprising soon \\
SynTF & 100 & 1 & good-for-naught yield rank impart motion-picture\_show manner\_of\_walking crepuscule at\_peace font flow asleep minute\_of\_arc finish rattling real bore non wish go\_on next \\
SynTF & 100 & 5 & gloomy throw military\_rating consecrate picture walk\_of\_life declension numb causa strike\_down numb moment end rattling very drill non worry materialize succeeding \\
SynTF & 100 & 10 & sad hold blackleg cave\_in flick base\_on\_balls decline asleep caseful settle benumbed second cease genuinely genuinely bore non deal pass\_off following \\
TEM & 100 & 1 & vandal kikkis bijomaru moliere betteralso crandall inmatesbr rightbr humped xvichiasergo midsummer keyboardist ferry mainlybecause crony sarlac overriding grossout drummed flopbecause \\
TEM & 100 & 5 & sorry gave rating give movie walk fall asleep case fell asleep minute end really really bored not caring happened next \\
TEM & 100 & 10 & sorry gave rating give movie walk fall asleep case fell asleep minute end really really bored not caring happened next \\
Mahalanobis & 100 & 1 & interpol disposable stuffing withstand cyclonic bardot dah ira hah unforced oom budget pulp saucepan simmer data srebrenica dna semitism diced \\
Mahalanobis & 100 & 5 & harry beheading restrictive und loos mal demolition subsistence proxy mountain blocking sung violently receiver concept diol buckingham leporidae written language \\
Mahalanobis & 100 & 10 & forgot opening rating give movie walk fall asleep attorney fell asleep minute ceiling code really deliriously story harrowing happened search \\
SanText & 100 & 1 & sorry ferro rating give truffaut walk fall asleep case fell asleep minute end really deutschland bored not afro paved next \\
SanText & 100 & 5 & sorry unconquerable rating conducting movie walk fall asleep concludes fell asleep minute end really musing bored not caring surly next \\
SanText & 100 & 10 & thankful this rating give upcoming lined fall asleep case slipped woke minute end really really bored not caring why next \\
Gumbel & 100 & 1 & thank gave disapproval take blockbuster sit rise woke trial slid woke minute start maybe really jaded so nurturing happened week \\
Gumbel & 100 & 5 & glad gave rating give hollywood walk end drunk case slipped asleep equalizer time everyone thought bored not sick happened day \\
Gumbel & 100 & 10 & ok brought disapproval come drama walked coming crawled complaint tumbled crawled kick start something think bored would caring knew start \\
Vickrey & 100 & 1 & constrict stir firefight jc elimination exhibitor ali birdman kissed imtiaz clothing gymnastics mile inexhaustible traveler meh decency mil album neagle \\
Vickrey & 100 & 5 & happy possession rating right tinge ball colombian toddler suburb fell gawk heartedly host guy fascinating bad escape truer ya unrealized \\
Vickrey & 100 & 10 & frankly later nielsen grab soundtrack walking since squirming relation gained wandered rush gap rely something downright believe widowed everybody time \\
CMP & 300 & 1 & gr gft expectable chakra grandparent gored magritte noo sniper breakfast meh substantive paternal verifiably viking flute erm striker muddies shane \\
CMP & 300 & 5 & relay neighborhood crime dubai hiroshima vampyre sandal kilcher estimate evers studiously mib cowgirl puzzle coldest stv danube berkley pulitzer del \\
CMP & 300 & 10 & cardiff interpretation efficiency hollywood movie stooped reacting sleep case export asleep minute dillon unable lot bored interfere trainor depressive gunter \\
SynTF & 300 & 1 & meritless devote rat ease\_up flick walkway fall departed character diminish asleep mo cease actually in\_truth blase not deal materialize future \\
SynTF & 300 & 5 & disconsolate afford rat devote movie base\_on\_balls downfall gone vitrine fell departed narrow terminal in\_truth truly tire non lovingness take\_place future \\
SynTF & 300 & 10 & bad cave\_in blackleg ease\_up movie paseo hang at\_peace font precipitate asleep mo stop truly actually bore not manage happen succeeding \\
TEM & 300 & 1 & peckenpahs urchin iñárritu lansburys streaming clout goosebump kissed welcomed maggies iwhippedi waswell occuped damme unbielevable calligraphy cameraman nula sharie british \\
TEM & 300 & 5 & sorry gave rating give movie walk fall asleep case fell asleep minute end really really bored not caring happened next \\
TEM & 300 & 10 & sorry gave rating give movie walk fall asleep case fell asleep minute end really really bored not caring happened next \\
Mahalanobis & 300 & 1 & perspicacious peace gf pellet gomer gargan raspberry kursk no prime wisconsin pickier reddin salvific designer clunkers incursion martyr hurd umm \\
Mahalanobis & 300 & 5 & sincere ha hrithik goalkeeper batman innes pole astral bellucci visa disfigured clan wale geometry faceoff simon sharia humperdink von faulty \\
Mahalanobis & 300 & 10 & sad summary rating age movie mum monarch asleep psychiatric fell asleep goalkeeper talker silver prototype improvising office caring thigpen declined \\
SanText & 300 & 1 & sorry gave rating give granddaughter walk fall asleep case fell asleep minute end really workhorse generate agreeing caring happened next \\
SanText & 300 & 5 & sorry gave rating give movie walk fall asleep hypothetically huggable selina memorably end crucially really starstruck insisting caring happened next \\
SanText & 300 & 10 & sorry gave rating give movie begun fall asleep concluded fell asleep minute end really really bored be caring happened next \\
Gumbel & 300 & 1 & embarrassed give indicating gave filmed anyway though awake reason dropped woke minute end definitely certainly tired though evera happen week \\
Gumbel & 300 & 5 & disappointed put rating giving movie walking fall awake example falling asleep equalizer ended really know bored although elderly happened expected \\
Gumbel & 300 & 10 & ashamed giving rating given starred walked coming sleep example slid fortunately came however obviously certainly bored be loving happening take \\
Vickrey & 300 & 1 & dah mayoral herein wachowski address ee corneau blazing ketchup observatory curled verdi thematic zen materialises ishwar wrestlemania nicholsons sonja interference \\
Vickrey & 300 & 5 & pepe pota eavesdrops hatching stunt yeop traumatizing takoma detained factly picher hitch light englund encyclopedia glanced calcium ditzy pasta chromosome \\
Vickrey & 300 & 10 & miserable assertion plunging invocation jerker sabre competing appetizer homicide dated suspended sanchez levy go consistency scene entertained flawed dreamt cbs \\
\bottomrule
\end{tabular}
}
\caption{Example text output on the IMDb dataset.}
\label{tab:ex1}
\end{table}

\begin{table}[htbp]
\resizebox{0.99\linewidth}{!}{
\begin{tabular}{r|llll}
\toprule
\multicolumn{1}{r|}{\textbf{Original:}} & \multicolumn{3}{c}{\textbf{Olympic Baseball Makes a Hit in Greece ATHENS (Reuters) - The team that did near the worst in Olympic baseball drew the most fans while the team among the most feared were not there}} \\ \hline
\multicolumn{1}{r|}{\textbf{Preproc:}} & \multicolumn{3}{c}{\textbf{olympic baseball make hit greece athens reuters team near worst olympic baseball drew fan team among feared}} \\ \hline
\multicolumn{1}{c}{\textbf{Mechanism}} & \multicolumn{1}{c}{\textbf{\textit{d}}} & \multicolumn{1}{c}{\textbf{$\epsilon$}} & \multicolumn{1}{c}{Sentence} \\
\midrule
\midrule
CMP & 50 & 1 & kazakhstani billboard physically humor debian pour mcconnaughey steller excavation orthodox moosehead fax forehand maggette republican hristo hornet \\
CMP & 50 & 5 & olympics mlb expensive smash jerez quote reuters valuable nablus wreaked olympics legion character era buyout mountainous believe \\
CMP & 50 & 10 & olympic baseball whatever hit greece athens reuters team northern worst olympic baseball stinging fan team among ridden \\
SynTF & 50 & 1 & Olympian baseball urinate pip Hellenic\_Republic Athinai reuters team\_up virtually worst Olympian baseball\_game pull devotee squad among fear \\
SynTF & 50 & 5 & Olympic baseball shuffling pip Ellas Greek\_capital reuters team close defective Olympic baseball thread buff squad among reverence \\
SynTF & 50 & 10 & Olympic baseball clear shoot Greece capital\_of\_Greece reuters squad virtually high-risk Olympic baseball thread devotee team among revere \\
TEM & 50 & 1 & cornerback mayer ampts trekker pariah rsen macrozonaris menu separately mae barren liven soreness assigning bayonne dts seat \\
TEM & 50 & 5 & olympic baseball make hit greece athens reuters team near worst olympic baseball drew fan team among feared \\
TEM & 50 & 10 & olympic baseball make hit greece athens reuters team near worst olympic baseball drew fan team among feared \\
Mahalanobis & 50 & 1 & st philippe rangana touchdown metway corrects mutate bernabeu deivi violence megabit liossia damp pheidippides maumoon disorder prehistoric \\
Mahalanobis & 50 & 5 & ioa hockey britain strip republic powerpoint mtfg evolution born hit oly specific inspirational equivalent human numerous outplays \\
Mahalanobis & 50 & 10 & olympic baseball make hit greece athens reuters team near worst olympic baseball drew drew team among causing \\
SanText & 50 & 1 & olympic baseball make hit follows athens reuters disposed near eserver amir baseball underscore fan team among untangle \\
SanText & 50 & 5 & olympic baseball make hit greece athens reuters team where dismal skating baseball drew fan team among feared \\
SanText & 50 & 10 & olympic baseball make hit greece athens reuters team near aftermath skating coaching drawing fan team among feared \\
Gumbel & 50 & 1 & olympic mlb could drove cyprus canberra posted team outside faced competition nba drawing buzz team well threatened \\
Gumbel & 50 & 5 & olympic yankee need another cyprus melbourne correspondent team northwest wake world nba responded joke squad represent blamed \\
Gumbel & 50 & 10 & champion sox making hit greece athens reporting team outside hardest olympic sox drawn sensation league especially feared \\
Vickrey & 50 & 1 & upload honorable mwai retake sternest xu amp trillion nazi bruin hayat publisher hitler quarterback mazda westbound amazon \\
Vickrey & 50 & 5 & berth baseball sheer storm hiptop gymnastic analysis shield hospital complete musharraf financed slain frankfurt playoff hartford professional \\
Vickrey & 50 & 10 & olympics slugger easier hitting bulgaria athens quoting coaching southwest devastating olympics hockey mccarthy mighty squad well politician \\
CMP & 100 & 1 & ligament al vote domecq balco mathematical amf minority misdemeanor anbar damarcus gnome veterinary seasoning dilshan dugout khost \\
CMP & 100 & 5 & bafta yawkey told unflattering javier panathinaikos poll team burst cheater rogge colorado gamespot fan corrects openly eritrea \\
CMP & 100 & 10 & olympic baseball come hit greece athens reuters team near mexico decider baseball break contributing team among feared \\
SynTF & 100 & 1 & Olympic baseball\_game induce hit Ellas Athens reuters team come\_on high-risk Olympic baseball tie winnow team\_up among revere \\
SynTF & 100 & 5 & Olympian baseball\_game establish striking Hellenic\_Republic Athens reuters team\_up skinny sorry Olympic baseball\_game withdraw lover team among reverence \\
SynTF & 100 & 10 & Olympic baseball\_game wee striking Greece Athens reuters squad virtually regretful Olympian baseball draw\_and\_quarter winnow team among fear \\
TEM & 100 & 1 & relaxation winnipegthree bermittelt tremor ballet recouped diameter hualien belgrade flutie mom visit workplace flexitronics ad violently \\
TEM & 100 & 5 & olympic baseball make hit greece athens reuters team near worst olympic baseball drew fan team among feared \\
TEM & 100 & 10 & olympic baseball make hit greece athens reuters team near worst olympic baseball drew fan team among feared \\
Mahalanobis & 100 & 1 & estuary condor uh indignity average reactor heatwave de smoking mideast threw unbeaten chesnot assateague alitalia lethal vaccine \\
Mahalanobis & 100 & 5 & origin softball verbal andhra collider bayern bad raced container binge olympic haired orleans biker suburb including dam \\
Mahalanobis & 100 & 10 & medalist mlb reach hit greece athens reuters team near cisco olympic baseball outing happier team often hearing \\
SanText & 100 & 1 & olympic selfish make hit shatter athens flagged team near worst bennger murkowski drew fan team among morose \\
SanText & 100 & 5 & olympic baseball it willis greece athens reuters team near worst olympic walking pushing lightning latter among feared \\
SanText & 100 & 10 & olympic baseball make throw greece athens thomson coaching nearby worst olympic baseball drew fan team among feared \\
Gumbel & 100 & 1 & swimming softball put coming bulgaria athens reporting team near aftermath medalist league came fan team leading believed \\
Gumbel & 100 & 5 & olympics softball need went albania olympic wsj team near worst olympic mlb attention spectator player especially feared \\
Gumbel & 100 & 10 & olympics baseball making went iceland nagano reuters team along dismal olympics baseball turned buzz soccer many feared \\
Vickrey & 100 & 1 & seung arabiya malik najaf pope sistani crore douri taiex retina stimulant dictator hristo muqtada awami du accurately \\
Vickrey & 100 & 5 & medal union siebel drug stealth chapter giant roadside repairing sluggish olympic optioned uruguay garden amateur elderly jolted \\
Vickrey & 100 & 10 & marathon streak kind drove italy sydney bloomberg retired town disruption relay nhl matt villain squad involving connection \\
CMP & 300 & 1 & paavo antivirus petronas peirsol yielding sodomy topix axa scandal java animated narcotic clearinghouse transmeta sega bush bnk \\
CMP & 300 & 5 & herbold rockies rutherford contactless greece pension healthday nfl lunar easyjet micronesia midriff marvel fan nvidia disarming altoona \\
CMP & 300 & 10 & aid hitter make hit greece intl cote philadelphia shattering worst surpassed baseball merz brawl car muslim shunned \\
SynTF & 300 & 1 & Olympic baseball lay\_down gain Greece capital\_of\_Greece reuters team nearly unsound Olympian baseball sop\_up fan team\_up among fear \\
SynTF & 300 & 5 & Olympic baseball score smasher Hellenic\_Republic Athens reuters squad close uncollectible Olympian baseball John\_Drew lover team\_up among venerate \\
SynTF & 300 & 10 & Olympic baseball give bang Hellenic\_Republic capital\_of\_Greece reuters squad virtually sorry Olympic baseball\_game Drew rooter team\_up among reverence \\
TEM & 300 & 1 & furless resuming edgy eyepods kazmirs senseis approving ederal physical benefited impose oreilly kastors hobbyist coxnyse found berkleigh \\
TEM & 300 & 5 & olympic baseball make hit greece athens reuters team near worst olympic baseball drew fan team among feared \\
TEM & 300 & 10 & olympic baseball make hit greece athens reuters team near worst olympic baseball drew fan team among feared \\
Mahalanobis & 300 & 1 & heraldic torrential fingerprint severe tamoxifen newspoll racketeer varejao jemaah marwan mental platinum atlanta centerfold incurable grudzielanek odumbe \\
Mahalanobis & 300 & 5 & accreditation shutout scientific ibf cide ekimov anxiously alberta unhurt overcapacity millionth fab espn marrero pleaded trampled terminal \\
Mahalanobis & 300 & 10 & olympic favorite move calculate alertness victor google team near fracture fired obscenity catalanotto snapping bubble violence curt \\
SanText & 300 & 1 & olympic baseball make hit greece wheelchair reuters team near description olympic baseball drew fan concerned among toting \\
SanText & 300 & 5 & placing baseball make touched greece athens packer offseason near worst advocating mlb drew fan team among feared \\
SanText & 300 & 10 & nagano mlb willing hit greece alerted reuters fourteen near devastating olympic baseball drew fan coached among feared \\
Gumbel & 300 & 1 & nagano baseball making hit greece olympics reuters time near terrible olympic baseball drew natsha team others fearing \\
Gumbel & 300 & 5 & olympics football even hit bulgaria athens rgentina squad close seen olympics baseball drew fan football although feared \\
Gumbel & 300 & 10 & olympic hockey even followed greece greece thomson squad nearby worst event mlb drew fan finished also possibly \\
Vickrey & 300 & 1 & babson skywatchers bared topix belmarsh mont abg yngling beatify nonfarm shoaib square felipe hindu production expression truce \\
Vickrey & 300 & 5 & streamcast linkedin ggp naples bomb mcauliffe implant michigan prairie poorer daisuke amr empire apiece que garland seasonal \\
Vickrey & 300 & 10 & decathlon devote lend critically piraeus janeiro cameraman castro southeast worst nagano hockey drawn bookstore season surging distracted \\
\bottomrule
\end{tabular}
}
\caption{Example text output on the AG News dataset.}
\label{tab:ex2}
\end{table}

\end{document}